\documentclass[journal]{IEEEtran}
\usepackage{graphicx}
\usepackage{booktabs}
\usepackage{cite}
\usepackage{tabularx}
\usepackage{multirow}
\usepackage[colorlinks=true]{hyperref}
\usepackage{subfigure} 
\usepackage[cmex10]{amsmath}
\usepackage{array}
\usepackage{mdwmath}
\usepackage{mdwtab}
\usepackage{eqparbox}
\usepackage{url}
\usepackage{amssymb}

\begin{document}

\title{A survey of face recognition techniques \\ under occlusion}

\author{
\IEEEauthorblockN{Dan Zeng,
Raymond Veldhuis and Luuk Spreeuwers
}\\
\IEEEauthorblockA{University of Twente, The Netherlands\\
\{d.zeng,r.n.j.veldhuis,l.j.spreeuwers\}@utwente.nl}
}
\maketitle
\begin{abstract}
The limited capacity to recognize faces under occlusions is a long-standing problem that presents a unique challenge for face recognition systems and even for humans. The problem regarding occlusion is less covered by research when compared to other challenges such as pose variation, different expressions, etc. Nevertheless, occluded face recognition is imperative to exploit the full potential of face recognition for real-world applications. In this paper, we restrict the scope to occluded face recognition. First, we explore what the occlusion problem is and what inherent difficulties can arise. As a part of this review, we introduce face detection under occlusion, a preliminary step in face recognition. Second, we present how existing face recognition methods cope with the occlusion problem and classify them into three categories, which are 1) occlusion robust feature extraction approaches, 2) occlusion aware face recognition approaches, and 3) occlusion recovery based face recognition approaches. Furthermore, we analyze the motivations, innovations, pros and cons, and the performance of representative approaches for comparison. Finally, future challenges and method trends of occluded face recognition are thoroughly discussed.
\end{abstract}



%
\IEEEpeerreviewmaketitle


\section{Introduction}

\IEEEPARstart{F}{ace} recognition is a computer vision task that has been extensively studied for several decades~\cite{best2018longitudinal}. Compared with other popular biometrics such as fingerprint, iris, palm, and vein, the face has a significantly better potential to recognize the identity in a non-intrusive manner. Therefore, face recognition is widely used in many application domains such as surveillance, forensics, and border control. With the development of deep learning techniques~\cite{krizhevsky2012imagenet,simonyan2014very,szegedy2015going,he2016deep,wen2016discriminative,liu2016large,szegedy2017inception,liu2017sphereface,wang2018cosface} and the publicly available large-scale face datasets~\cite{bansal2017umdfaces,yi2014learning,parkhi2015deep,cao2018vggface2,nech2017level,guo2016ms}, face recognition performance has improved substantially~\cite{learned2016labeled,trigueros2018face}. However, face recognition systems still tend to perform far from satisfactory when encountering challenges such as large-pose variation, varying illumination, low resolution, different facial expressions, and occlusion. Generally, face images stored in a gallery are of high quality and free from the above degradations, while probe faces are suffering from what can be seen as \textit{a missing data problem} due to these challenges. Consequently, fewer facial parts are available for recognition, which induces a mismatch between feature available in probe faces and gallery faces.

\begin{figure}[t]
\centering
\includegraphics[width=3.2in]{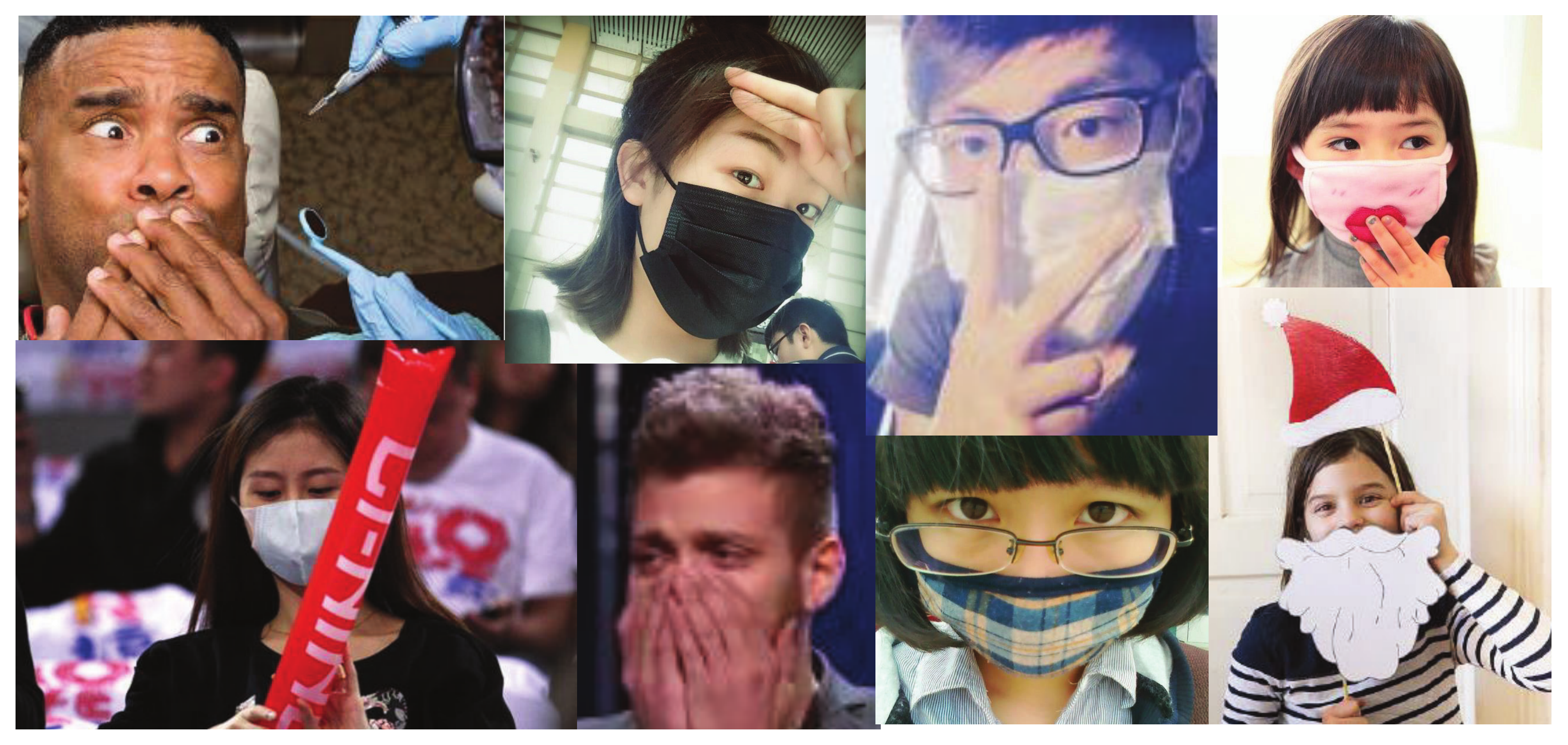}
\caption{Examples of occluded face images from the MAFA dataset.}
\label{fig:occlusion_examples}
\end{figure}

\begin{table}[t]
\renewcommand\arraystretch{1.1}
\caption{A categorization of Occlusion Challenges}\label{tab:occlusion_challenges}
\centering
\begin{tabularx}{0.5\textwidth}{|l|X|}
\hline
\textbf{Occlusion Scenario} & \textbf{Examples} \\
\hline
Facial accessories & eyeglasses, sunglasses, scarves, mask, hat, hair\\
External occlusions & occluded by hands and random objects\\
Limited field of view & partial faces\\
Self-occlusions& non-frontal pose\\
Extreme illumination& part of face highlighted\\
\hline
\multirow{3}{*}{Artificial Occlusions}&random black rectangles\\
&random white rectangles\\
&random salt \& pepper noise\\
\hline
\end{tabularx}
\end{table}
Facial occlusion~\cite{ekenel2009facial,oloyede2017evaluating} is considered one of the most intractable problems because we do not have prior knowledge about the occluded part, which can be anywhere and of any size or shape in a face image. From a practical point of view, it is not feasible to collect a large-scale training dataset with all possible occlusions in a realistic scenario to use deep learning techniques. Therefore, the problem of face recognition under occlusions remains a challenge. Facial occlusion occurs when the subject wears accessories such as a scarf, a face mask, glasses, a hat, etc., or when random objects are present in front of the face. The recognition accuracy has been compromised in some way because of the higher inter-class similarity and the more considerable intra-class variations caused by occlusion. Facial appearance changes substantially due to occlusion, as illustrated in Fig.~\ref{fig:occlusion_examples}. We present a categorization of occlusion challenges in different scenarios with their typical occlusion examples (see Table~\ref{tab:occlusion_challenges}). Pose variation can partially be seen as a self-occlusion problem caused by a large rotation of the head. The self-occlusion problem due to pose variation is usually dealt with in pose correction and therefore not discussed here.

{In most cases, occluded face recognition~(OFR) involves querying a gallery consisting of occlusion-free faces using a probe image from an alternative test dataset of occluded faces. Occluded faces rely on either the collection of real occlusions or synthetic occlusions. We first break down OFR research scenarios in the most obvious way by the pairs of images considered. Fig.~\ref{fig:OFR_problems} offers an illustration of the five categories regarding OFR testing scenarios. More specifically, five widely used testing scenarios for OFR, ranging from most real to least real, are:
\begin{itemize}
  \item \textbf{Real occlusions}: gallery images are mugshots free from occlusion while probe images are faces occluded by realistic images such as sunglasses, or a scarf.
  \item \textbf{Partial faces}: gallery images are mugshots free from occlusion while test face images are partial faces; hence the name partial face recognition is given by researchers.
  \item \textbf{Synthetic occlusions}: gallery images are faces in the wild which are captured from uncontrolled scenarios while probe faces are blocked with synthetic occlusions to simulate real occlusions. 
  \item \textbf{Occluding rectangle}: gallery images are occlusion-free mugshots while test face images are occluded with a rectangle such as white and black rectangles.
  \item \textbf{Occluding unrelated images}: gallery images are mugshots free from occlusion while test face images are occluded with unrelated images such as a baboon, or a non-square image.
\end{itemize}

\begin{figure}[h!]
\centering
\includegraphics[width=0.45\textwidth]{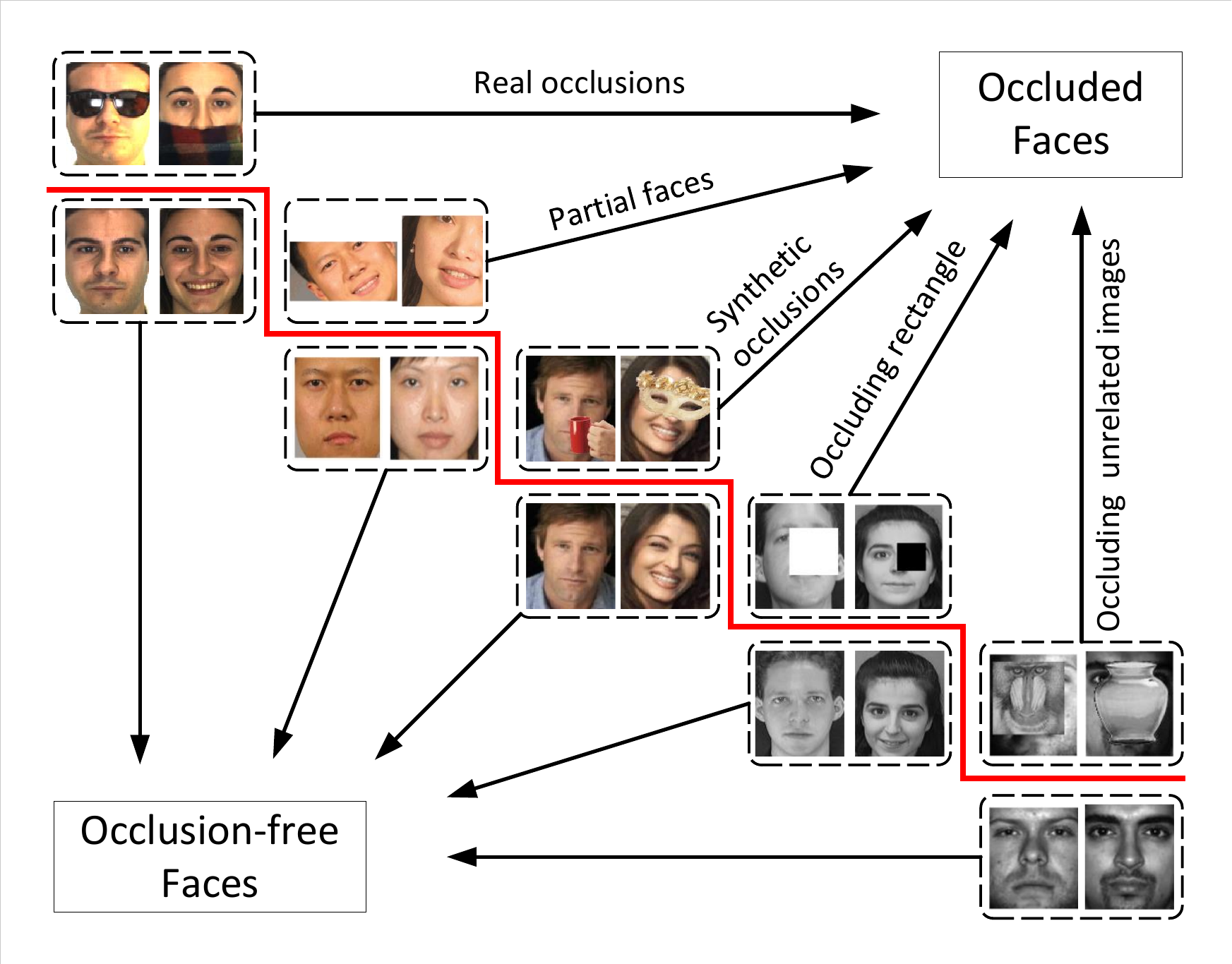}
\caption{Different occluded face recognition testing scenarios involved in OFR.}
\label{fig:OFR_problems}
\end{figure}
}

\begin{figure}[h!]
\centering
\includegraphics[width=0.5\textwidth]{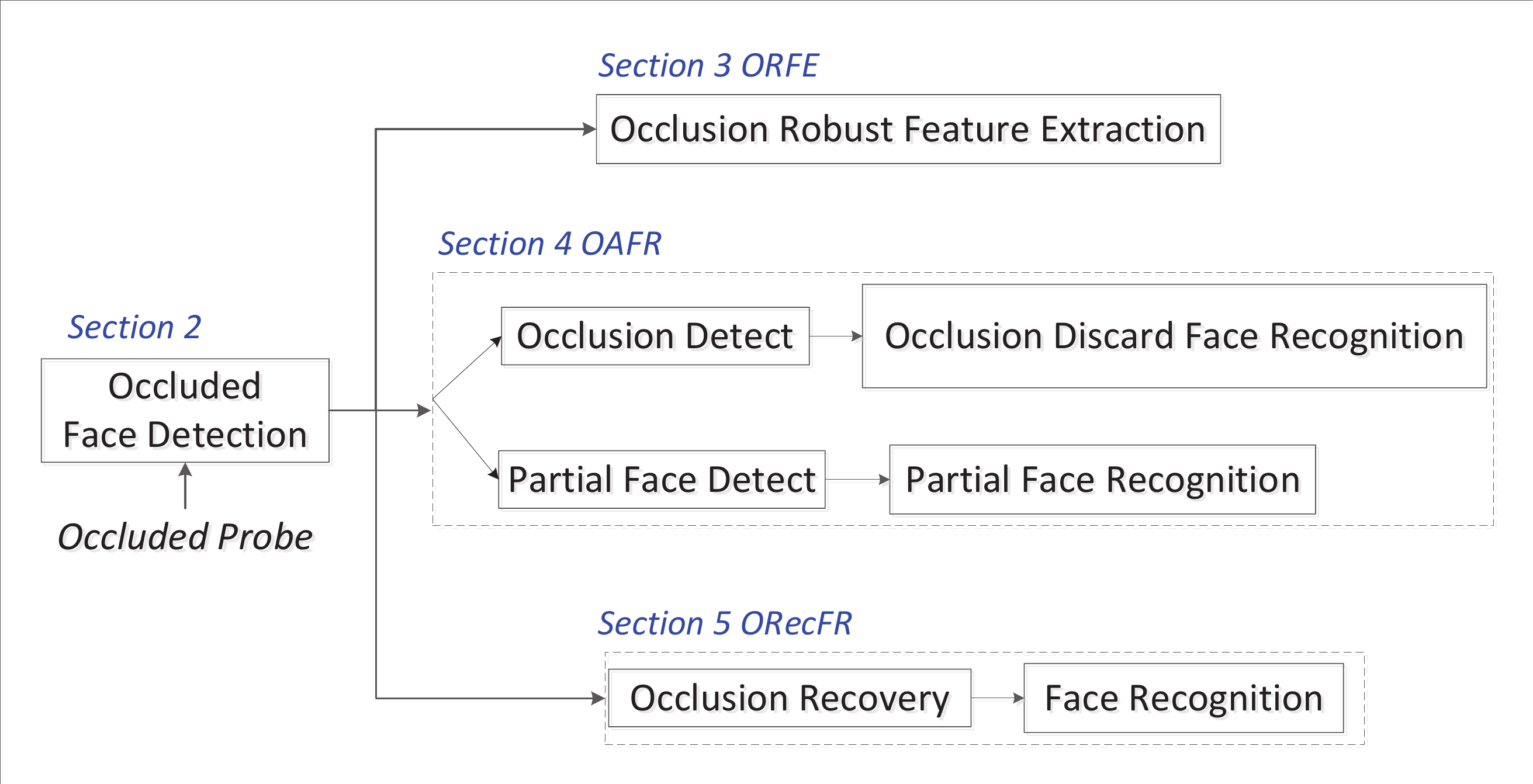}
\caption{The three categories of methods for face recognition under occlusion challenges.}
\label{fig:methods_category}
\end{figure}

Approaches to recognizing faces under occlusions can be broadly classified into three categories~(shown in Fig.~\ref{fig:methods_category}), which are 1) occlusion robust feature extraction~(ORFE), 2) occlusion aware face recognition~(OAFR) and 3) occlusion recovery based face recognition~(ORecFR). An OFR system consists of three components, each corresponding to an important design decision: cross-occlusion strategy, feature extraction, and comparison strategy. Of these components, the second and third have analogues in general face recognition, while cross-occlusion strategy is unique to OFR.
\begin{itemize}
\item \textbf{ORFE category} searches for a feature space that is less affected by facial occlusions. Generally, patch-based engineered and learning-based features are used as the cross-occlusion strategy.
\item \textbf{OAFR category} is explicitly aware where the occlusion is. Generally, occlusion-discard is applied as the cross-occlusion strategy. As a result, only visible face parts qualify for face recognition~(i.e., feature extraction, feature comparison). Furthermore, we classify partial face recognition approaches as OAFR because they exclude occlusion parts from face recognition, assuming that visible parts are ready in the beginning.
\item \textbf{ORecFR category} intends to recover an occlusion-free face from the occluded face to meet the demands of conventional face recognition systems. In other words, it takes occlusion recovery as the cross-occlusion strategy.
\end{itemize}

Numerous methods have been proposed to push the frontier of face recognition research. Several comprehensive surveys~\cite{azeem2014survey,lahasan2017survey,dagnes2018occlusion} have been published for face recognition under occlusion. Of the existing works, the survey by Lahasan et al.~\cite{lahasan2017survey} on face recognition under occlusion challenges presents a thorough overview of the approaches before 2017, which is the most relevant to this paper. However, there are at least two reasons why a new survey on occluded face recognition is needed. \textit{First, the explosive growth of face recognition techniques these years has stimulated many innovative contributions to handle occluded face recognition problems.} The increased number of publications over the last few years calls for a new survey for occluded face recognition, including up-to-date approaches, especially deep learning techniques. \textit{Second, several large-scale occluded face datasets have become publicly available in recent years.} Without large-scale training data of occluded face images, deep learning models cannot function well~\cite{mehdipour2016comprehensive}. Recently, the MAFA dataset~\cite{ge2017detecting} is accessible for occluded face detection, and the IJB-C dataset~\cite{maze2018iarpa} is introduced as a general evaluation benchmark to include meta-information regarding the occlusion~(i.e., occlusion location, occlusion degree). Predictably, these datasets would encourage occluded face recognition to develop faster. The proposed survey provides a systematic categorization of methods for face recognition. Specifically, occluded face detection techniques are briefly reviewed because an OFR system requires the application of occluded face detection as the first step. Moreover, newly published and innovative papers addressing occlusion problems are thoroughly reviewed. Finally, we represent comparative performance evaluations in terms of occluded face detection and face recognition on widely used datasets as well as newly-developed large-scale datasets.

The remainder of the paper is organized as follows: occluded face detection techniques are introduced in Section 2. Methods of occlusion robust feature extraction are described and analyzed in Section 3. We review occlusion-aware face recognition approaches in Section 4. Then Section 5 briefly studies occlusion-recovery face recognition methods. A performance evaluation of the reviewed approaches is given in Section 6. In section 7, we discuss future challenges to datasets as well as to research. Finally, we draw an overall conclusion for occluded face recognition.




\section{Occluded face detection}
\begin{figure}[h!]
\centering
\includegraphics[width=0.5\textwidth]{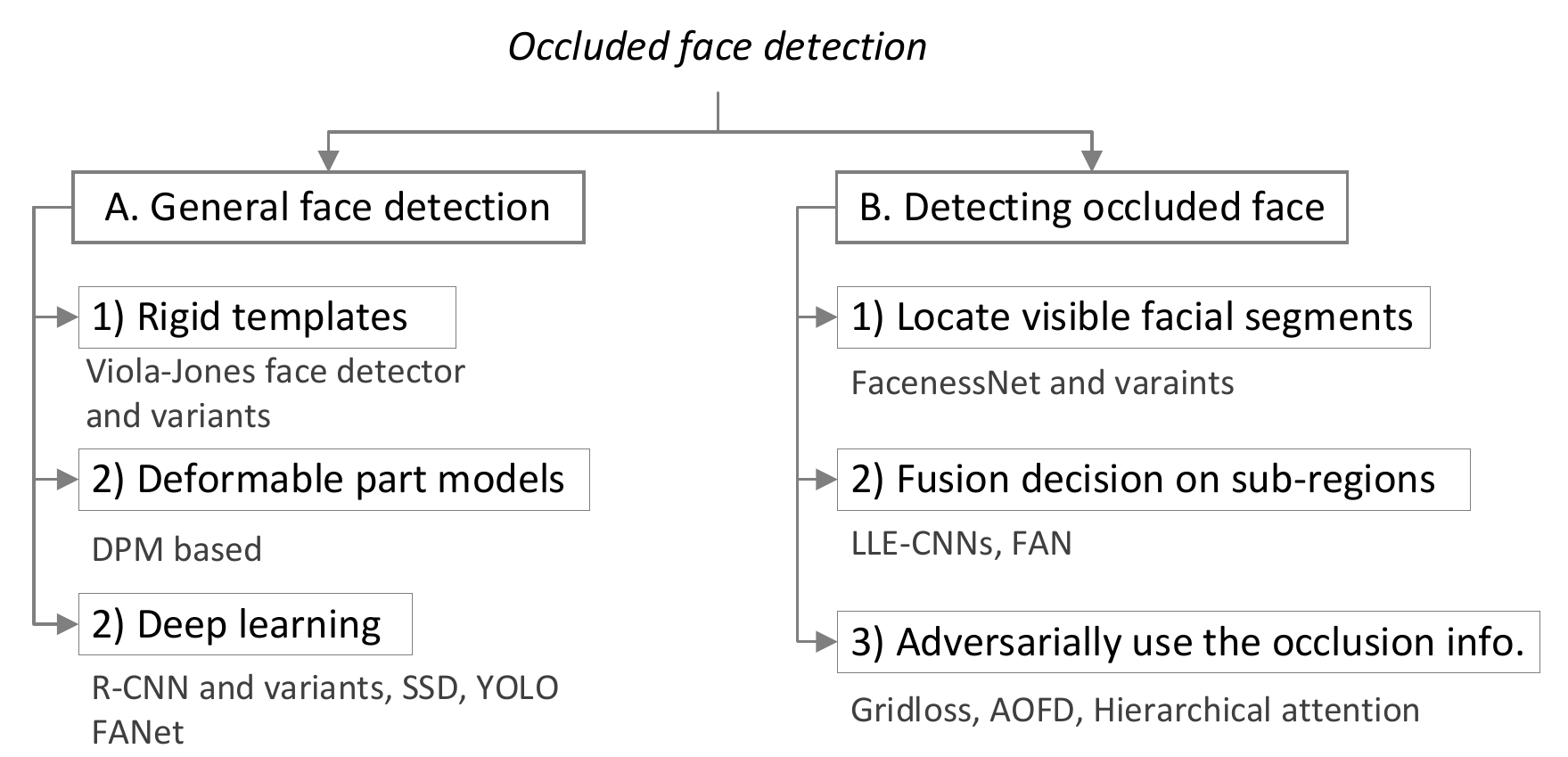}
\caption{Methods used in occluded face detection.}
\label{fig:ofd}
\end{figure}

In this section, we break down occluded face detection into two parts containing~\textit{general face detection} methods which can be applied to detect occluded face, and~\textit{occluded face detection} methods which are designed specifically to tackle the occlusion issue in face detection. As for general face detection, we briefly study relevant methods for the simplicity. We then elaborate occluded face detection methods that are specifically designed to detect occluded faces. One way to classify the methods can be seen in Fig.~\ref{fig:ofd}.

\subsection{General face detection} 
Face detection generally intends to detect the face that is captured in an unconstrained environment. It is challenging due to large-pose variation, varying illumination, low resolution, and occlusion etc.~\cite{zafeiriou2015survey}. Approaches to general face detection can roughly be classified into three categories, which are 
\begin{itemize}
\item Rigid templates based face detection.
\item Deformable part models~(DPM) based face detection.
\item Deep convolutional neural networks~(DCNNs) based face detection. 
\end{itemize}

The Viola-Jones face detector and its variations~\cite{viola2001rapid,viola2004robust} are typical in the rigid templates based category, which utilizes Haar-like features and AdaBoost to train cascaded classifiers and can achieve good performance with real-time efficiency. However, the performance of these methods can drop dramatically when handling real-world application~\cite{yang2014aggregate}. In contrast, DPM based face detection can achieve significantly better performance based on the cost of high computational complexity~\cite{yan2014fastest}. A third, most promising category of research is DCNNs based face detection~\cite{li2015convolutional,chen2014joint,zhang2014improving,zhang2016joint,jiang2017face,sun2018face,zhang2017s3fd,yang2016wider,bai2018finding,li2019pyramidbox++,zhang2020robust}. Some methods~\cite{chen2014joint,zhang2014improving,zhang2016joint} joint face detection with face alignment to exploit their inherent correlation to boost the performance. There are two major branches of object detection framework: (i)~region proposals based CNN~(i.e.,two-stage detectors), such as R-CNN~\cite{girshick2014rich}, fast R-CNN~\cite{girshick2015fast}, faster R-CNN~\cite{ren2015faster}; (ii)~region proposals free CNN~(i.e.,one-stage detectors), such as the Single-Shot Multibox Detector~(SSD)~\cite{liu2016ssd}, YOLO~\cite{redmon2016you}. In short, two-stage detectors achieve higher performance but are time-consuming. One-stage detectors have significant computational advantages but compensate by less accurate detection results. Some methods~\cite{jiang2017face,sun2018face,zhang2017s3fd,zhang2020feature} treat a face as a natural object and adopt techniques from object detection in face detection. Most recently, finding tiny faces has become popular in face detection and superior performance has been achieved~\cite{hu2017finding,bai2018finding,li2019pyramidbox++,zhang2020robust}. 

Recent years have witnessed promising results of exploring DCNNs for face detection with the introduction of Widerface~\cite{yang2016wider}, which offers a wide pose variation, significant scale difference~(tiny face), expression variation, make-up, severe illumination, and occlusion. Up to now, Feature Agglomeration Networks~(FANet)~\cite{zhang2020feature}, a single-stage face detector, achieves the state-of-art performance on several face detection benchmarks include the FDDB~\cite{jain2010fddb}, the PASCAL Face~\cite{yan2014face}, and the Widerface benchmark~\cite{yang2016wider}. To exploit inherent multi-scale features of a single convolutional neural network, FANet introduced an Agglomeration Connection module to enhance the context-aware features and augment low-level feature maps with a hierarchical structure so that it can cope with scale variance in face detection effectively. Besides, Hierarchical Loss is proposed to train FANet to become stable and better in an end-to-end way. \textbf{In short, methods that achieve remarkable detection performance, for example on the Widerface dataset, also provide a solid solution for occluded face detection.}

\subsection{Detecting occluded face}

Detecting partially occluded faces aims to locate the face region in a given image where occlusion is present. Handling occlusion in  face detection is challenging due to the unknown location and the type of occlusions~\cite{chen2006modification}. Recently, occluded face detection is beginning to attract the attention of researchers; therefore a few publications are reviewed. At the same time, detecting occluded pedestrians is a long-standing research topic that has been intensively studied during the past few decades. Therefore, many researchers borrow techniques from pedestrian detection~\cite{zhang2018occlusion,zhou2018bi,guo2018occlusion} to push the frontier of occluded face detection by treating occlusion as the dominating challenge during the detection. \textbf{Most occluded face detection methods report their performance on the MAFA dataset~\cite{ge2017detecting} while general face detection methods do not, which means it is not a level playing field for general face detection and occluded face detection.} Approaches to detect partially occluded faces are roughly clustered as 1)~locating visible facial segments to estimate a full face; 2)~fusing the detection results obtained from face sub-regions to mitigate the negative impact of occlusion; 3)~using the occlusion information to help face detection in an adversarial way.

\subsubsection{Locating visible facial segments to estimate face}

If visible parts of a face are known, then difficulties in face detection due to occlusions are largely relieved. Observing that facial attributes are closely related to facial parts, the attribute-aware CNNs method~\cite{yang2015facial} intends to exploit the inherent correlation between a facial attribute and visible facial parts. Specifically, it discovers facial part responses and scores these facial parts for face detection by the spatial structure and arrangement. A set of attribute-aware CNNs are trained with specific part-level facial attributes (e.g., mouth attributes such as big lips, open mouth, smiling, wearing lipstick) to generate facial response maps. Next, a scoring mechanism is proposed to compute the degree of face likeliness by analyzing their spatial arrangement. Finally, face classification and bounding box regression are jointly trained with the face proposals, resulting in precise face locations. The results on FDDB~\cite{jain2010fddb}, PASCAL Face~\cite{yan2014face} and AFW~\cite{zhu2012face} demonstrate that the proposed method is capable of yielding good performance. In particular, they can achieve a high recall rate of $90.99\%$ on FDDB. In short, Ref.~\cite{yang2015facial} is the first systematic study to attempt face detection with severe occlusion without using realistic occluded faces for training.

More recently, the extension faceness-net~\cite{yang2018faceness} improves the robustness of feature representations by involving a more effective design of CNN. As a result, it has achieved compelling results on the Widerface dataset~\cite{yang2016wider}, which is challenging in terms of severe occlusion and unconstrained pose variations. However, it requires the use of labeled attributes of facial data to train attribute-aware CNN, which impairs its practical use in some way. 

\subsubsection{Fusing detection results obtained from face sub-regions}
In paper~\cite{mahbub2016partial}, a facial segment based face detection technique is proposed for mobile phone authentication with faces captured from the front-facing camera. The detectors are AdaBoost cascade classifiers trained with a local binary pattern (LBP) representation of face images. They train fourteen segments-based face detectors to help cluster segments in order to estimate a full face or partially visible face. As a result, this method could achieve excellent performance on the Active Authentication Dataset (AA-01)~\cite{samangouei2015attribute,zhang2015touch}. However, the use of simple architecture increases speed by compromising detection accuracy.

The introduction of MAFA~\cite{ge2017detecting} offers plenty of faces wearing various masks, which  contributes significantly to the occluded face detection, especially of masked faces. Based on this dataset, an LLE-CNNs~\cite{ge2017detecting} is proposed to benchmark the performance of masked face detection on the MAFA dataset. They extract candidate face regions with high-dimensional descriptors by pre-trained CNNs and employ locally linear embedding (LLE) to turn them into similarity-based descriptors. Finally, they jointly train the classification and regression tasks with CNNs to identify candidate facial regions and refine their position.

To avoid high false positives due to masks and sunglasses, a face attention network (FAN) detector~\cite{wang2017face} is proposed to highlight the features from the face region. More specifically, the FAN detector integrates an anchor-level attention mechanism into a single-stage object detector like Feature Pyramid Networks~\cite{lin2017feature}. The attention supervision information is obtained by filling the ground-truth box and is associated with the ground-truth faces which match the anchors at the current layer. The attention maps are first fed into an exponential operation and then combined with feature maps. As a result, the method is capable of achieving impressive results on the Widerface~\cite{yang2016wider} with an $88.5\%$ average precision on the hard subset as well as on an $88.3\%$ average precision on MAFA~\cite{ge2017detecting} dataset.

\subsubsection{Occlusion information adversarially used for detection}

Apart from selecting the visible facial parts and fusing results obtained from face sub-regions, it is a third way to minimize the adverse effects of face detection due to occlusions. One promising approach is to use a novel grid loss~\cite{opitz2016grid}, which has been incorporated into the convolutional neural network to handle partial occlusion in face detection. It is based on the observation that partial occlusions would confuse a subset of detectors, whereas the remaining ones can still make correct predictions. To this end, this work regards occluded face detection as a particular single-class object detection problem, inspired by other works on object detection~\cite{farfade2015multi,hosang2015taking,li2015convolutional,sermanet2013pedestrian}. Furthermore, the proposed grid loss minimizes the error rate on face sub-blocks independently rather than over the whole face to mitigate the adverse effect of partial occlusions and to observe improved face detection accuracy. 

Using the occluded area as an auxiliary rather than a hindrance is a feasible solution to help face detection adversely. Adversarial occlusion-aware face detection (AOFD)~\cite{chen2017masquer} is proposed to detect occluded faces and segment the occlusion area simultaneously. They integrate a masking strategy into AOFD to mimic different occlusion situations. More specifically, a mask generator is designed to mask the distinctive part of a face in a training set, forcing the detector to learn what is possibly a face in an adversarial way. Besides, an occlusion segmentation branch is introduced to help detect incomplete faces. The proposed multitask training method showed superior performance on general as well as masked face detection benchmarks. To cope with different poses, scales, illumination, and occlusions, Wu et al.~\cite{wu2019hierarchical} introduce a hierarchical attention mechanism, applying long short-term memory~(LSTM) to predict face-specific attention. In this way, it can further model relations between the local parts and adjust their contribution to face detection. The proposed method achieves promising performance compared with Faster R-CNN~\cite{ren2015faster}.



\section{Occlusion robust feature extraction}

\begin{figure}[h!]
\centering
\includegraphics[width=0.5\textwidth]{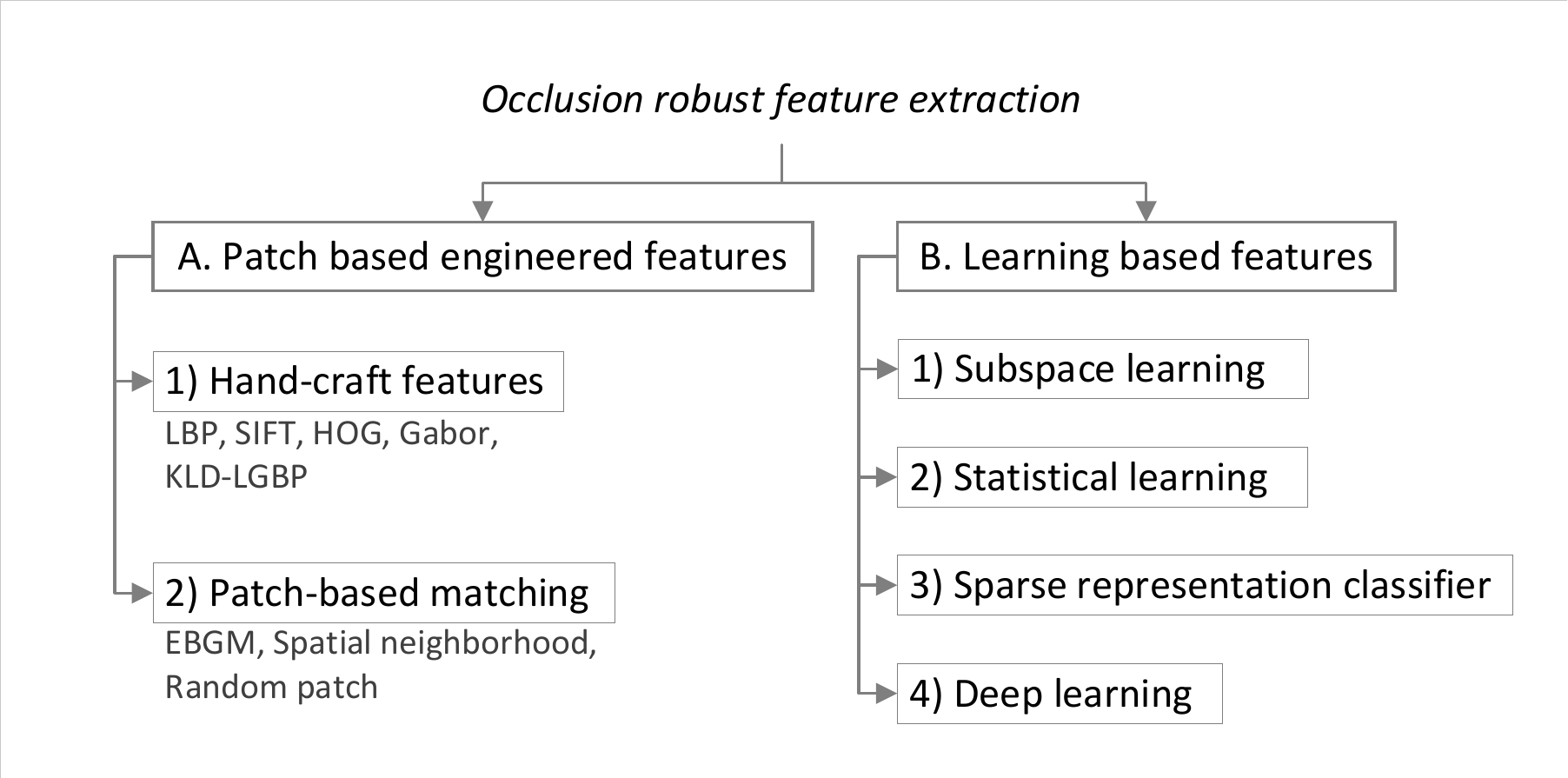}
\caption{Methods used in occlusion robust feature extraction approaches.}
\label{fig:orfe}
\end{figure}

If the extracted features are reasonably robust to the occlusion, then difficulties in face recognition due to occlusion are relieved. The aim is to extract features that are less affected by occlusions~(outliers) while preserving the discriminative capability. We group the approaches into engineered features and learning-based features. The former generally extract handcraft features from explicitly defined facial regions, which do not require optimization or a learning stage. The latter extract features by using learning-based methods such as linear subspace methods, sparse representation classification, or nonlinear deep learning techniques. One way to classify the methods can be seen in Fig.~\ref{fig:orfe}.

\subsection{Patch-based engineered features}
Facial descriptors obtained in an engineered way are rather efficient because they can: (i)~be easily extracted from the raw face images; (ii)~discriminate different individuals while tolerating large variability in facial appearances to some extent; (iii)~lie in a low feature space so as to avoid a computationally expensive classifier. Generally, engineered features~(i.e., handcraft features) are extracted from facial patches and then concatenated to represent a face. Therefore, a fusion strategy can be imported to reduce the adverse effects of occluded patches in some way. Alternatively, patch-based matching can be used for feature selection to preserve occlusion-free discriminative information.

These methods in general require precise registration such as alignment based on eye coordinates for frontal faces and integrate the decisions from local patches to obtain a final decision for face recognition. This is problematic because these methods rely on robust face alignment under occlusion, but the eyes are likely to be occluded. In short, these approaches are not realistic for application since most often the face images need to be aligned well to facilitate feature extraction of meaningful facial structure. 

\subsubsection{Handcraft features}
Local Binary Patterns~(LBP)~\cite{ahonen2004face,ahonen2006face} is used to derive a novel and efficient facial image representation and has been widely used in various applications. LBP and variants~\cite{zhao2007dynamic,liao2007learning} retain popularity and succeed so far in producing good results in biometrics, especially face recognition. The main idea is to divide the face image into multiple regions, from which to extract LBP feature distributions independently. These descriptors are then concatenated to form an enhanced global descriptor of the face. For distance measurement between two faces, weighted Chi square distance is applied, accounting for some facial features being more important in human face recognition than others. The Scale Invariant Feature Transform~(SIFT) descriptor~\cite{lowe1999object} is popular in object recognition and baseline matching and can also be applied to face recognition~\cite{zhang2009faceprint}. SIFT is largely invariant to changes in scale, translation, rotation, and is also less affected by illumination changes, affine or 3D projection. Similarly to SIFT, the Histograms of Oriented Gradient~(HOG) descriptor~\cite{dalal2005histograms} has been proposed to handle human detection and has been extended to cope with object detection as well as visual recognition. The main idea is to characterize local object appearance and shape with the distribution of local integrity gradients. After applying a dense~(in fact, overlapping) grid of HOG descriptors to the detection window, the descriptors are combined to suit the further classifier. Contrary to the integrity oriented methods, Gabor filters and other frequency oriented approaches construct the face feature from filter responses. Generally, the filter responses computed for various frequencies and orientations from a single or multiple spatial locations are combined to form the Gabor feature~\cite{zou2007comparative}. Phase information instead of magnitude information from Gabor features contains discrimination and is thus widely used for recognition~\cite{zhang2006histogram}.
Features based on Gabor filters are versatile. By post-processing they can be converted, for example, to binary descriptors of texture similar to LBPs. Ref.~\cite{zhang2007local} proposes KLD-LGBP, in which Kullback-Leibler Divergence (KLD) is introduced to measure the distance between the local Gabor binary patterns~(LGBP) feature~\cite{zhang2005local} of the local region of test images and that of the unoccluded local region of reference faces. They define the probability of occlusions of that area as the distance between two distributions of local regions and further use it as the weight of the local region for the final feature matching. The main drawback of this method is the high dimensionality of LGBP features, which are the combination of Gabor transform, LBP, and a local region histogram on local face regions.

\subsubsection{Patch-based matching}
Beside representation, the distance metric also plays an important role. Elastic and partial matching schemes bring in a lot of flexibility when handling challenges in face recognition. Elastic Bunch Graph Matching~(EBGM)~\cite{wiskott1997face} uses a graph to represent a face, each node of the graph corresponding to Gabor jets extracted from facial landmarks. The matching method is used to calculate the distance between corresponding representations of two faces. To take advantage of elastic and partial matching, Ref.~\cite{hua2009robust} proposes a robust matching metric to match Difference of Gaussian~(DoG) filter descriptor of a facial part against its spatial neighborhood in the other faces and select the minimal distance for face recognition. Specifically, they extract $N$ local descriptors from densely overlapped image patches. During the matching, each descriptor in one face is picked up to match its spatial neighborhood descriptors and then the minimal distance is selected, which is effective in reducing the adverse effects of occlusion. A random sampling patch-based method~\cite{cheheb2017random} has been proposed to use all face patches equally to reduce the effects of occlusion. It trains multiple support vector machine~(SVM) classifiers with selected patches at random. Finally, the results from each classifier are combined to enhance the recognition accuracy. Similarly to elastic bunch graph matching~\cite{wiskott1997face}, dynamic graph representation~\cite{ren2019dynamic} is proposed to build feature graphs based on node representations. Each node corresponds to certain regions of the face image and the edge between nodes represents the spatial relationship between two regions. Dynamic graph representation can enable elastic and partial matching, where each node in one face image is matched with adjacent nodes in the other face image for face recognition, which brings in robustness to occlusion especially when encountering partially occluded biometrics.

\subsection{Learning-based features}


Compared with engineered features, learned features are more flexible when various occlusion types at different locations are present. Features learned from training data can be effective and have potentially high discriminative power for face recognition. Unlike regular images, face images share common constraints, such as containing a smooth surface and regular texture. Face images are in fact confined to a face subspace. Therefore, \textit{subspace learning} methods have been successfully applied to learn a subspace that can preserve variations of face manifolds necessary to discriminate among individuals. Taking the occlusion challenge as a major concern, it is natural to apply \textit{statistical methods} on face patches allowing for the fact that not all types of occlusion have the same probability of occurring.~\textit{Sparse representation classifier} methods, which fully explore the discriminative power of sparse representation and represent a face with a combination coefficient of training samples, have been the mainstream approach to handle various challenges in face recognition for a long time. The last few years have witnessed a great success of~\textit{deep learning} techniques~\cite{masi2018deep}, especially deep convolutional neural networks~(DCNN) for uncontrolled face recognition applications.

\subsubsection{Subspace learning}
Approaches such as the Principal component analysis~(PCA)~\cite{turk1991eigenfaces} and the Linear Discriminant Analysis~(LDA)~\cite{belhumeur1997eigenfaces} are the two most representative methods in learning the subspace for feature extraction. Eigenface~\cite{turk1991eigenfaces} uses PCA to learn the most representative subspace from training face images. Fisherface~\cite{belhumeur1997eigenfaces} uses LDA to explore a discriminant subspace that differentiates faces of different individuals by incorporating the class labels of faces images. One limitation for these appearance-based methods is that proper alignment specifically based on the eye location is required. Modular PCA~\cite{gottumukkal2004improved} builds multiple eigen spaces (eigenfeatures) in the regions of facial components (e.g., eyes, nose, and mouth) to achieve variance tolerance, especially for face variances. Ref.~\cite{de2003framework} uses an influence function technique to develop robust subspace learning for a variety of linear models within a continuous optimization framework. Ref.~\cite{leonardis2000robust} uses the property of PCA to detect ``outliers'' and occlusions, and calculates the coefficients using the rest of the pixels. Inspired by robust PCA~\cite{leonardis2000robust}, a new approach combining discriminative power and reconstructive property has been proposed by Fidler et al.~\cite{fidler2006combining}, which can achieve good classification results and have the construction abilities to detect outliers (occlusions). Specifically, object recognition and face recognition applications are used to estimate the robustness of LDA coefficients. Besides, objects’ orientation application, a regression task, is introduced to estimate the robustness of extended CCA coefficients. Ref.~\cite{kim2005effective} proposes independent component analysis~(ICA) to use locally salient information from important facial parts for face recognition. This part-based representation method could achieve better performance in case of partial occlusions and local distortions than PCA and LDA. Unlike linear subspace methods, nonlinear subspace methods use nonlinear transforms to convert a face image into a discriminative feature vector, which may attain highly accurate recognition in practical scenarios. Ref.~\cite{savvides2006partial} proposes kernel correlation feature analysis (KCFA) on facial regions, i.e., eye-region, nose-region, and mouth-region to cope with partial face recognition. The proposed KCFA for feature extraction outperforms the linear approaches PCA~\cite{turk1991eigenfaces}, LDA~\cite{belhumeur1997eigenfaces} and variants of kernel methods~\cite{zheng2004real,yang2002kernel}.

\subsubsection{Statistical learning}
In real-world applications, not all types of occlusions have the same probability of occurring; for example, a scarf and sunglasses often have a higher probability of occurrence compared with others. Hence, it is natural to apply statistical learning on face patches to account for their occlusion possibility. One early work in this direction is Ref.~\cite{martinez2002recognizing}, which takes a probabilistic approach, i.e., a mixture of Gaussians to compensate for partially occluded faces. Specifically, they analyze local regions divided from a face in isolation and apply the probabilistic approach to find the best match so that the recognition system is less sensitive to occlusion. Ref.~\cite{tan2005recognizing} extends the work by representing the face subspace with self-organizing maps (SOM). It presents the similarity relationship of the subblocks in the input space in the SOM topological space. Furthermore, they use the soft k nearest neighbor ensemble classifier to efficiently capture the similarity relationship obtained by SOM projections, which in turn improves the whole system's robustness. However, this method assumes knowledge of the occluded parts in advance. Since partial occlusion affects only specific local features, Ref.~\cite{hotta2008robust} proposes a local Gaussian kernel for feature extraction and trains the SVM using the summation of local kernels as combine features. However, it is suboptimal to use the same size local kernel rather than select the appropriate size of the local kernel for local feature extraction. In Ref.~\cite{tan2009face}, a non-metric partial similarity measure based on local features is introduced to capture the prominent partial similarity among face images while excluding the unreliable and unimportant features. 

In paper~\cite{seo2011robust}, a face recognition method is proposed that takes partial occlusions into account by using statistical learning of local features. To this end, they estimated the probability density of the SIFT feature descriptors observed in training images based on a simple Gaussian model. In the classification stage, the estimated probability density is used to measure the importance of each local feature of test images by defining a weighted distance measure between two images. Based on this work, they extended the statistical learning based on local features to a general framework~\cite{seo2014robust}, which combines the learned weight of local features and feature-based similarity to define the distance measurement. However, feature extraction from the local region cannot code the spatial information of faces. Besides, the unreliable features from the occluded area are also integrated into the final representation, which will reduce the performance. McLaughlin et al.~\cite{mclaughlin2017largest} try to solve random partial occlusion in test images using the largest matching areas at each point on the face. They assume that the occluded test image region can be modeled by an unseen-data likelihood with a low posterior probability. More specifically, they de-emphasize the local facial area with low posterior probabilities from the overall score for each face and select only reliable areas for recognition, which results in improved robustness to partial occlusion.

\subsubsection{Sparse representation classifier}

Apart from these statistical learning methods, several algorithms use sparse representation classifiers~(SRC) to tackle the occlusion in face recognition. Ever since its introduction~\cite{wright2009robust}, SRC attracts increasing attention from researchers. This method explores the discriminative power of sparse representation of a test face. It uses a linear combination of training samples plus sparse errors to account for occlusion or corruption as its representation. Yang et al.~\cite{yang2013gabor} propose a Gabor feature based Robust Representation and Classification (GRRC) scheme. They use Gabor features instead of pixel values to represent face images, which can increase the ability to discriminate identity. Moreover, the use of a compact Gabor occlusion dictionary requires less expensive computation to code the occlusion portions compared with that of the original SRC. To investigate the effectiveness of the proposed method, they conduct extensive experiments to recognize faces with block occlusions as well as real occlusions. A subset of the AR database was used in this experiment. It consists of 799 images (about eight samples per subject) of non-occluded frontal views with various facial expressions for training. The sunglasses test set contains $100$ images with the subject wearing sunglasses~(with a neural expression), and the scarves test set contains $100$ images with the subject wearing a scarf~(with a neural expression). The proposed GRRC achieves $93.0\%$ recognition accuracy on sunglasses and $79\%$ accuracy on scarves, which outperforms the original SRC by $5\%$ and $20\%$, respectively. In paper~\cite{liu2016face}, artificial occlusions are included to construct training data for training sparse and dense hybrid representation framework. The results show that artificially introduced occlusions are important to obtain discriminative features. Structured occlusion coding~(SOC)~\cite{wen2016structured} is proposed to employ an occlusion-appended dictionary to simultaneously separate the occlusion and classify the face. In this case, with the use of corresponding parts of the dictionary, face and occlusion can be represented respectively, making it possible to handle large occlusion, like a scarf. In paper~\cite{fu2017efficient}, efficient locality-constrained occlusion coding~(ELOC) is proposed to greatly reduce the running time without sacrificing too much accuracy, inspired by the observation that it is possible to estimate the occlusion using identity-unrelated samples.

Recently, another work~\cite{yang2017joint} attempts face recognition with single sample per person and intends to achieve robustness and effectiveness for complex facial variations such as occlusions. It proposes a joint and collaborative representation with local adaptive convolution feature~(ACF), containing local high-level features from local regular regions. The joint and collaborative representation framework requires ACFs extracted from different local areas to have similar coefficients regarding a representation dictionary. Ref.~\cite{gao2017learning} proposes to learn a robust and discriminative low-rank representation~(RDLRR) for optimal classification, which aiming to prepare the learned representations optimally for classification as well as to capture the global structure of data and the holistic structure of each occlusion induced error image. The results demonstrate that the method can yield better performance in case of illumination changes, real occlusion as well as block occlusion.
Ref.~\cite{li2017ubiquitous} combines center-symmetric local binary patterns~(CSLBP) with enhanced center-symmetric local binary patterns~(ECSLBP) to build the SRC dictionary for occluded face recognition. In Ref.~\cite{yu2017discriminative}, discriminative multi-scale sparse coding~(DMSC) is proposed to address single-sample face recognition with different kinds of occlusion. More specifically, it learns the auxiliary dictionary to model the possible occlusion variations from external data based on PCA and proposes a multi-scale error measurement strategy to detect and disregard outlier pixels due to occlusion. Ref~\cite{wu2018occluded} proposes a hierarchical sparse and low-rank regression model and uses features based on image gradient direction, leading to a weak low-rankness optimization problem. The model is suited for occluded face recognition and yields better recognition accuracy. In another work, NNAODL~(nuclear norm based adapted occlusion dictionary learning)~\cite{du2019nuclear} has been proposed to construct corrupted regions and non-corrupted regions for occluded face recognition. The same occlusion parts in training images are used to construct the occlusions while normal training images are used to reconstruct non-occlusion regions, leading to improved computing efficiency. To cope with occluded face recognition with limited training samples, Ref.~\cite{zheng2020novel} proposes a structural element feature extraction method to capture the local and contextual information inspired by the human optic nerve characteristics for face recognition. Besides, an adaptive fusion method is proposed to use multiple features consisting of a structural element feature, and a connected-granule labeling feature. It applies Reinforced Centrosymmetric Local Binary Pattern~(RCSLBP) to better handle the robustness to the occlusion. Finally, few-shot sparse representation learning is applied for few-shot occluded face recognition.

\subsubsection{Deep learning}
Face representation obtained by DCNN is vastly superior to other learning-based methods in discriminative power, owing to the use of massive training sets~\cite{masi2018deep}. Face
verification performance has been boosted due to advanced
deep CNN architectures~\cite{krizhevsky2012imagenet,simonyan2014very,he2016deep,szegedy2015going,szegedy2017inception} and the development
of loss functions~\cite{wen2016discriminative,liu2016large,liu2017sphereface,wang2018additive,deng2019arcface,wang2018cosface,schroff2015facenet,sun2014deep}. From a practical point of view, we can obtain occlusion-robust face representation if a massive training dataset is provided to contain sufficient occluded faces for a deep network to learn to handle occlusions~\cite{zhou2015naive}. However, it is tough to collect such a dataset. Data augmentation seems to be a plausible solution to obtain a large number of occluded faces. Training with augmented occluded faces ensures the features are extracted more locally and equally~\cite{osherov2017increasing}. Lv et al.~\cite{lv2017data} synthesize occluded faces with various hairstyles and glasses for data augmentation to enlarge the training dataset, which alleviates the impact of occlusions. Specifically, $87$ common hairstyle templates with various bangs and $100$ glasses templates are collected to synthesize training face images with different hairstyles and face images with different glasses, to enable the CNN model is robust to various hairstyles and glasses. The method indeed relieves the data deficiency problem and results in improved performance. However, its use is limited to handling sunglasses and hair in recognition and does not solve the occlusion problem in general. In paper~\cite{trigueros2018enhancing}, instead of using synthetic occluded faces directly, they first identify the importance of face regions in an occlusion sensitivity experiment and then train a CNN with identified face regions covered to reduce the model's reliance on these regions. Specifically, they propose to augment the training set with face images with occlusions located in high-effect regions~(the central part of the face) more frequently than in low effect regions~(the outer parts of the face). This forces the model to learn more discriminative features from the outer parts of the face, resulting in less degradation when the central part is occluded. However, pre-defined occlusions may cause performance degradation when dealing with face images with an occlusion that is not of the same size. Cen et al.~\cite{cen2019dictionary} propose a DDRC~(deep dictionary representation based classification) scheme to alleviate the occlusion effect in face recognition, where the dictionary is used to linearly code the deep features that are extracted from a convolutional neural network. The dictionary is composed of deep features of the training samples and an auxiliary dictionary associated with the occlusion patterns of the testing face samples. However, the proposed DDRC assumes that the occlusion pattern of the test faces is included in the auxiliary dictionary, which limits its use.

\section{Occlusion aware face recognition}
\begin{figure}[h!]
\centering
\includegraphics[width=0.5\textwidth]{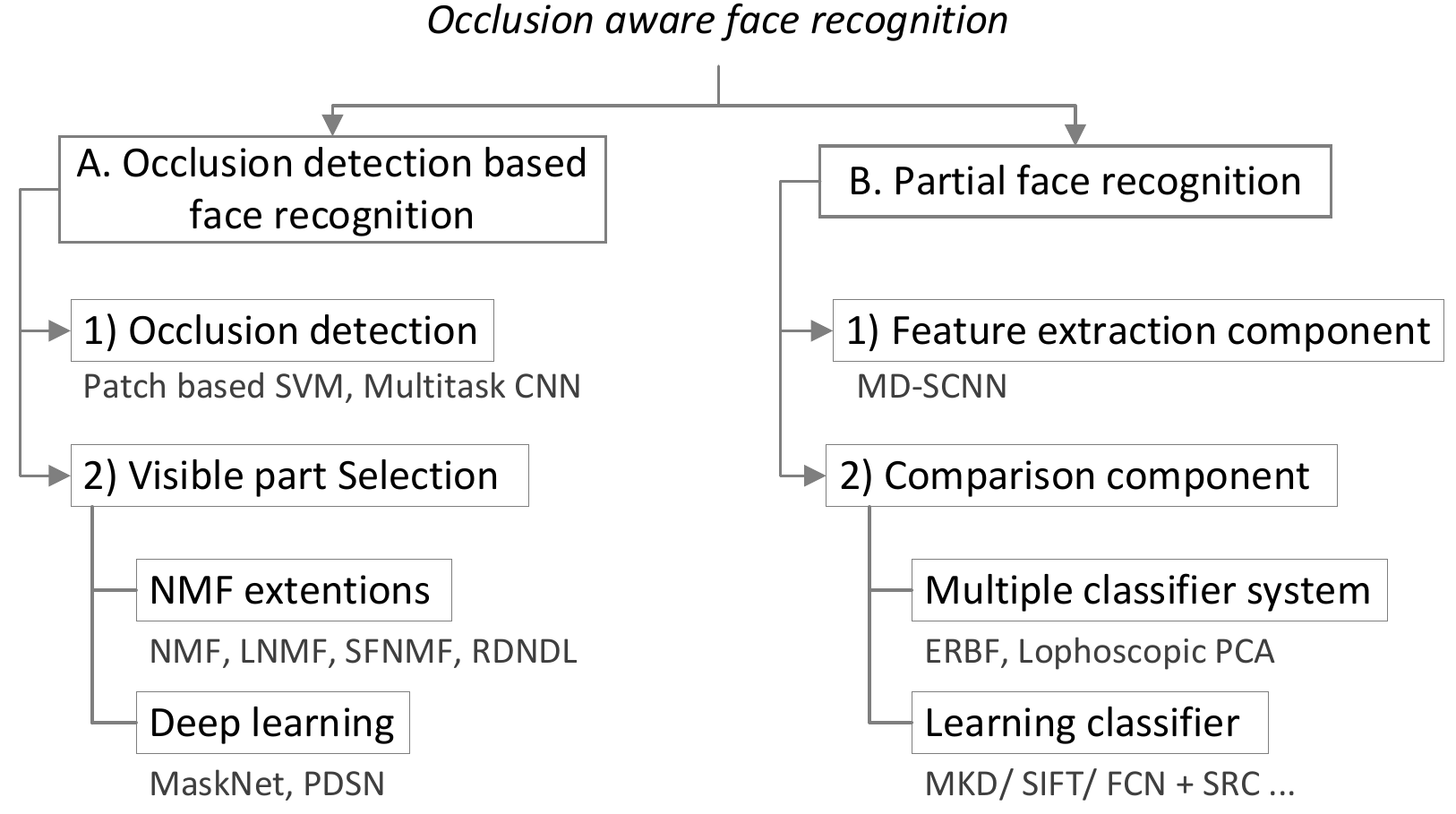}\label{oafr}
\caption{Method classification used in occlusion aware face recognition approaches.}
\label{fig:oafr}
\end{figure}

If only visible facial parts are used for recognition, then the occlusion problem is mitigated to some extent. These approaches that explicitly exclude the occlusion area are called occlusion aware face recognition (OAFR). There are two groups of methods that constitute OAFR. One is occlusion detection based face recognition, which detects the occlusions first and then obtains a representation from the non-occluded parts only. The other one is partial face recognition, which assumes that a partial face is available and aims to use a partial face for recognition. Occlusion detection is ignored during partial face recognition. A taxonomy of occlusion aware face recognition is shown in Fig.~\ref{fig:oafr}.


\subsection{Occlusion detection based face recognition}
To explicitly make use of facial parts for face recognition, some methods explicitly detect the occlusion and perform face recognition based on their results. The other techniques obtain visible facial parts for face recognition based on the prior knowledge of occlusion, which is called visible part selection.

\subsubsection{Occlusion detection} An intuitive idea to deal with occlusions in face recognition is to detect the occlusions first and then recognize the face based on unoccluded facial parts. Methods use predefined occlusion types as a substitute for arbitrary occlusion in different locations to simplify the occlusion challenge. Usually, scarves and sunglasses are used as representative occlusions because of their high probability of appearance in the real world. Based on this idea, Ref.~\cite{oh2008occlusion} introduces a 1-NN~(nearest neighbor) threshold classifier to detect the occluded face region in a PCA subspace learned from training data of occlusion-free patches. Then they apply the selective local non-negative matrix factorization method to select features corresponding to occlusion-free regions for recognition.
Some early works~\cite{chen2011occluded,min2011improving} employ a binary classifier to search for the occluded area and incorporate only the unoccluded parts for comparison. Specifically, they first divide the face into multiple non-overlapping regions and then train an SVM classifier to identify whether the facial patch is occluded or not. By excluding occluded regions, improved overall recognition accuracy is observed. However, the performance is far from satisfactory and very sensitive to the training dataset. Ref.~\cite{andres2014face} proposes to utilize compressed sensing for occlusion detection by using occluded and occlusion-free faces of the same identity. For the recognition process, discriminative information is extracted by excluding the occluded areas detected.

Since occlusions can corrupt the features of an entire image in some way, deep learning techniques are developed to alleviate the problem by producing a better representation. Ref.~\cite{xia2015face} designs a convolutional neural network to detect the occlusion in a multitask setting. More specifically, there are four region-specific tasks for occlusion detection, and each aims to predict the occlusion probability of the specific component: left eye, right eye, nose, and mouth. However, predicting only predefined occlusions limits flexibility, and inaccuracy of occlusion detection can, in return, harm the recognition performance. 

\subsubsection{Visible part selection} Some works select visible facial parts for recognition and skip occlusion detection by assuming the prior knowledge of occlusion. Ref.~\cite{neo2010development} compares several subspace-based methods including PCA, Non-negative matrix factorization~(NMF)~\cite{lee1999learning}, Local NMF~\cite{li2001learning} and Spatially Confined NMF~\cite{neo2005novel} and uses the partial information available for face recognition. During face recognition, the eye region is selected when people are wearing masks or veils, and the bottom region is used when people are wearing glasses. This method has a deficiency in flexibility use because well-aligned predefined subregions are hard to obtain in the real scenario. A paper~\cite{ou2018robust} in this direction extends NMF to include adaptive occlusion estimation based on the reconstruction errors. Low-dimensional representations are learned to ensure that features of the same class are close to that of the mean class center.  This method does not require prior knowledge of occlusions and can handle large continuous occlusions.

In paper~\cite{wan2017occlusion}, a proposed MaskNet is added to the middle layer of CNN models, aiming to learn image features with high fidelity and to ignore those distorted by occlusions. MaskNet is defined as a shallow convolutional network, which is expected to assign lower weights to hidden units activated by the occluded facial areas. Recently, Song et al.~\cite{song2019occlusion} propose a pairwise differential siamese network~(PDSN) to build the correspondence between the occluded facial block and the corrupted feature elements with a mask learning strategy. The results show improved performance on synthesized and realistic occluded face datasets.


\subsection{Partial face recognition}

It is worth mentioning that we classify partial face recognition as occlusion aware methods because partial face recognition skips the occlusion detection phase and focuses on face recognition when arbitrary patches are available, which can be seen as implicit occlusion awareness. Partial faces frequently appear in unconstrained scenarios, with images captured by surveillance cameras or handheld devices (e.g., mobile phones) in particular. To the best of our knowledge, research for partial face detection has so far been ignored in literature reviews. It is essential to search for the semantic correspondence between the partial face~(arbitrary patch) and the entire gallery face since it is meaningless to compare the features of different semantic facial parts. The semantic correspondence can be completed either in the feature extraction phase to extract invariant and discriminative face features, or in the comparison phase to construct a robust face classifier. Feature extraction and comparison methods can be developed to address the partial face recognition problem. Therefore, we categorize the methods as feature-aware and comparison-aware methods.

\subsubsection{Feature extraction component} As for feature-aware methods, Multiscale Double Supervision Convolutional Neural Network~(MDSCNN)~\cite{he2016multiscale} is proposed to have multiple networks trained with facial patches of different scales for feature extraction. Multiscale patches are cropped from the whole face and aligned based on their eye-corners. Each network is training with face patches of a scale. The weights of multiple networks are combined learned to generate final recognition accuracy. Even though this method can yield good feature representations, it is troublesome to train 55 different Double Supervision Convolutional Neural Networks~(DSCNNs) according to different scaled patches. It is time-consuming in practice because window sliding is needed to generate multiscale patches for recognition.

\subsubsection{Comparison component}
Comparison-aware methods facilitate the semantic correspondence in the comparison phase of face recognition. Among the comparison based approaches, the multiple classifier systems use a voting mechanism to tolerate the misalignment problem to an extent. In this regard, Gutta et al.~\cite{gutta2002investigation} present an Ensemble of Radial Basis Function~(ERBF) Networks consisting of nine RBF components, and each of which is determined according to the number of clusters and overlap factors. A verification decision is based on the output generated by the RBFs. In paper~\cite{tarres2005novel}, they propose the Lophoscopic PCA method to recognize faces in the absence of part of the relevant information. For this purpose, masks (black rectangles) that cover specific regions of the face, including left eye, right eye, both eyes, nose, mouth, and no mask are introduced to compute different subspaces. They learn the weights for each subspace during training. When classifying the subject, weights regarding each face subspace are considered and combined for recognition. These methods are not entirely satisfactory but need improvement, particularly in terms of their recognition rate.

Alternatively, a learning-based classifier compensates for the difficulty in the alignment of partial face recognition. Ref.~\cite{liao2013partial} conducts the first systematic study to recognize an arbitrary patch of a face image without preliminary requirement on face alignment, which is regarded as a milestone in the field of partial face recognition. The proposed method employs the multi-keypoint descriptors~(MKD) to represent a holistic or partial face with a variable length. Descriptors from a large gallery construct the dictionary, making it possible to sparsely represent the descriptors of the partial probe image and infer the identity of the probe accordingly. However, SRC requires a sufficient number of faces to cover all possible variations of a person, which hinders the realization of a practical application.

Even though most learning-based classifier papers are SRC based, there is a small group that develops similarity measures to address the partial face recognition problem. Ref.~\cite{hu2013robust} utilizes the instance-to-class distance to address the partial face recognition with scale-invariant feature transform~(SIFT) descriptors. The similarity between each probe patch and gallery image is obtained by comparing a set of local descriptors of one probe image to the nearest neighbor descriptors of all gallery images with the sparse constraint. As an improvement, Ref.~\cite{weng2013robust} and~\cite{weng2016robust} consider partial face recognition as a feature set matching problem, where they explicitly use geometric features and textural features for simultaneous matching. Robust point set matching~(RPSM)~\cite{weng2016robust} considers both geometric distribution consistency and textural similarity. Moreover, a constraint on the affine transformation is applied to prevent unrealistic face warping. However, these methods would fail to work if face keypoints are unavailable due to occlusions. Moreover, the computation complexity is high, which makes the recognition process slow.

Recently, there is a trend to combine the deep learning and SRC methods to tackle the partial face recognition~\cite{he2018dynamic,he2019dynamic}. Dynamic feature learning~\cite{he2018dynamic} combines a fully convolutional network (FCN) with sparse representation classification (SRC) for partial face recognition. The sliding window matching proposed in paper~\cite{zheng2015partial} searches for the most similar gallery part by sliding the window of the same size as the partial probe. The combination of sliding window matching based on FCN and SRC brings a promising performance. As an improved version, Ref.~\cite{he2019dynamic} uses multi-scale representations in dynamic feature learning, which increase the ability to tolerate the misalignment between partial probe patches and the gallery face. 

\section{Occlusion recovery based face recognition}

\begin{figure}[h!]
\centering
\includegraphics[width=0.5\textwidth]{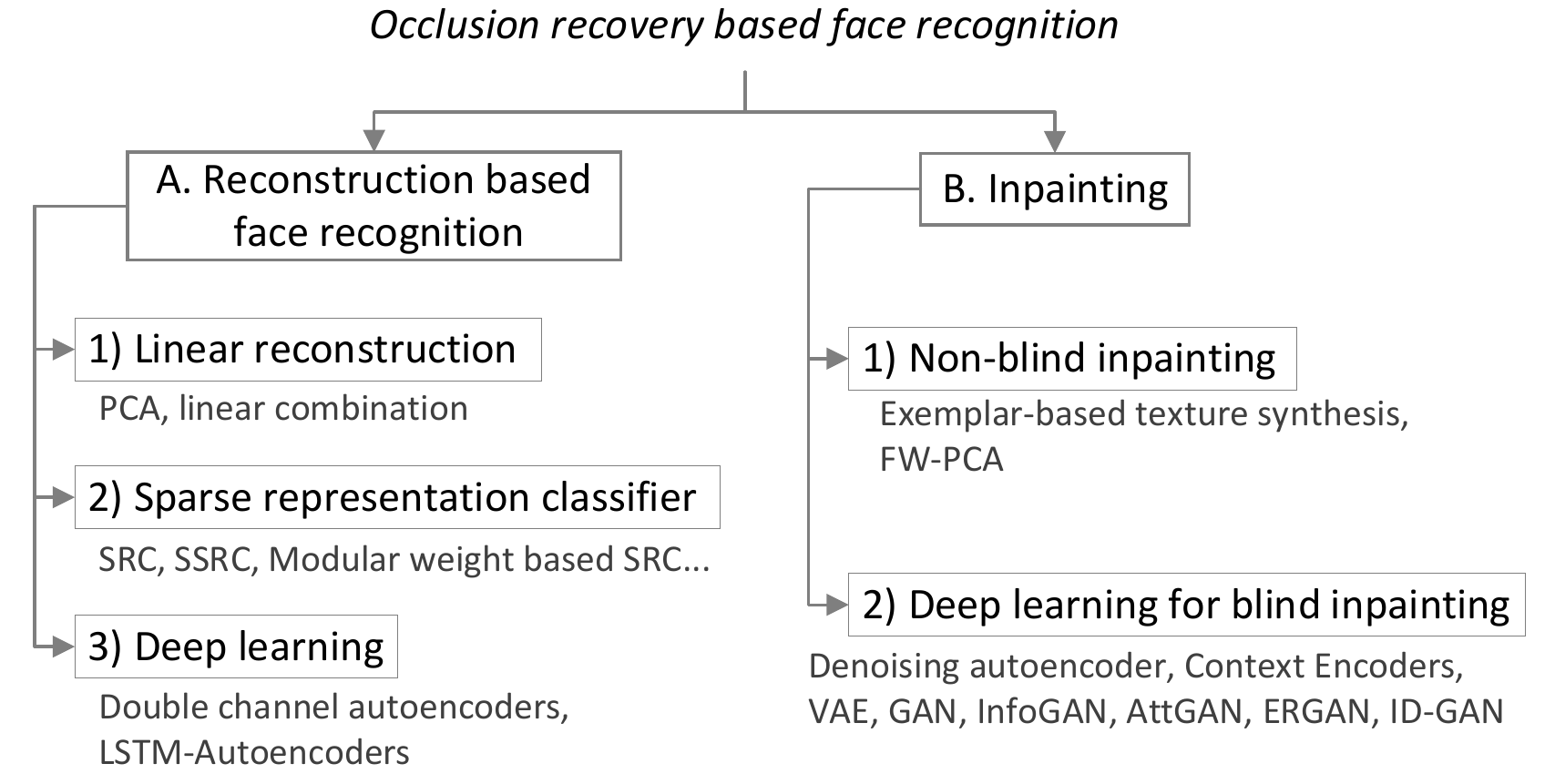}
\caption{Classification of methods used in occlusion recovery based face recognition approaches.}
\label{fig:orfr}
\end{figure}

Apart from addressing the occlusion problem in feature space, one intuitive idea is to take occluded face recovery as a substitution to solve the occlusion in image space. Occlusion recovery methods recover a whole face from the occluded face, which allows the direct application of conventional face recognition algorithms. Existing occlusion recovery methods for face recognition use (i)~reconstruction based techniques for face recognition, or (ii)~inpainting techniques, which treat the occluded face as an image repairing problem. A possible way to classify the methods can be seen in Fig.~\ref{fig:orfr}.

\subsection{Reconstruction based face recognition}
Image-based two-dimensional reconstructions carefully study the relationship between occluded faces and occlusion-free faces. The reconstruction techniques are classified as linear reconstruction, sparse representation classifier~(dictionary learning), and deep learning techniques.

\subsubsection{Linear reconstruction} As for reconstruction techniques, Ref.~\cite{park2005glasses} utilizes PCA reconstruction and recursive error compensation to remove the eye occlusions caused by glasses. It combines a Markov Random Field model with a parse representation of occluded faces to improve the reconstruction of corrupted facial regions. There are many variants~\cite{de2003framework,leonardis2000robust,fidler2006combining} employing PCA to detect outliers or occlusion and then reconstruct occlusion-free face images. Ref.~\cite{jia2008face} estimates an occluded test image as a linear combination of training samples of all classes, which allows having occlusions in training and testing sets. Distinct facial areas are weighted differently so that only non-occluded facial parts are used for reconstruction. 

\subsubsection{Sparse representation classifier} The sparse representation classifier~\cite{wright2009robust} is considered to be the pioneering work on occlusion robust face recognition. This method explores the discriminative power of sparse representation of a test face. It uses a linear combination of training samples plus sparse errors accounting for occlusions or corruption as its representation. To better tackle the occlusion, the SRC introduces an identity matrix as an occlusion dictionary on the assumption that the occlusion has a sparse representation in this dictionary. Ref.~\cite{zhou2009face} improves sparse representation by including prior knowledge of pixel error distribution. In paper~\cite{deng2011graph}, they design a graph model-based face completion system for partially occluded face reparation. They leverage image-based data mining to find the best-matched patch to guide occluded face synthesis from the images, derived from the selection of sparse representation classification (SRC). The final face completion proceeds in the graph-based domain with the help of graph Laplace. 

Similarly, Ref.~\cite{li2013reconstruction} proposes a reconstruction approach consisting of occlusion detection and face reconstruction. First, the downsampled SRC is used to locate all possible occlusions at a low computing complexity. Second, all discovered face pixels are imported into an overdetermined equation system to reconstruct an intact face.
An innovative solution for the occlusion challenge is presented by structured sparse representation based classification (SSRC)~\cite{ou2014robust} to learn an occlusion dictionary. The regularization term of mutual incoherence forces the resulting occlusion dictionary to be independent of the training samples. This method effectively decomposes the occluded face image into a sparse linear combination of the training sample dictionary and the occlusion dictionary. The recognition can be executed on the recovered occlusion-free face images. Nevertheless, this method requires retraining of the model to handle the different occlusions. 

In paper~\cite{zhao2015modular}, a new criterion to compute modular weight-based SRC is proposed to address the problem of occluded face recognition. They partition a face into small modules and learn the weight function according to the Fisher rate. The modular weight is used to lessen the effect of modules with low discriminant and to detect the occlusion module. More recently, Ref.~\cite{iliadis2017robust} proposes a robust representation to model contiguous block occlusions based on two characteristics. The first characteristic introduces a tailored potential loss function to fit the errors of distribution. Specifically, a Laplacian sparse error distribution or more general distributions based on M-Estimators. The second characteristic models the error image, which is the difference between the occluded test face and the unoccluded training face of the same identity, as low-rank structural. Wang et al.~\cite{wang2017varying} proposed a method equipped with two stages: varying occlusion detection stage consisting of occlusion detection and elimination, and iterative recovery stage consisting of occluded parts being recovered and unoccluded parts being reserved. With the use of iteratively recovered strategy, this joint occlusion detecting and recovery method can produce good global features to benefit classification.

\subsubsection{Deep learning} A few works use deep learning techniques for occlusion reconstruction. One is work~\cite{cheng2015robust}, which extends a stacked sparse denoising autoencoder to a double channel for facial occlusion removal. It adopts the layerwise algorithm to learn a representation so that the learned encoding parameters of clean data can transfer to noisy data. As a result, the decoding parameters are refined to obtain a noise-free output. The other work~\cite{zhao2018robust} proposes to combine the LSTM and autoencoder architectures to address the face de-occlusion problem. The proposed robust LSTM-Autoencoders (RLA) model consists of two LSTM components. One spatial LSTM network encodes face patches of different scales sequentially for robust occlusion encoding, and the other dual-channel LSTM network is used to decode the representation to reconstruct the face as well as to detect the occlusion. Additionally, adversarial CNNs are introduced to enhance the discriminative information in the recovered faces.

\subsection{Inpainting}
Image inpainting techniques are widely used to carefully obtain occlusion-free images and are not limited to face images. Inpainting techniques focus on repairing the occluded images and leave face recognition out of consideration. They can be divided into 1)~non-blind inpainting and 2)~blind inpainting categories, depending on whether the location information of corrupted pixels is provided or not. Deep learning is an effective approach to blind inpainting.

\subsubsection{Non-blind inpainting} Those techniques fill in the occluded part of an image using the pixels around the missing region. Exemplar-based techniques that cheaply and effectively generate new texture by sampling and copying color values from the source are widely used. In paper~\cite{criminisi2004region}, a non-blind inpainting method proposes a unified scheme to determine the fill order of the target region, using an exemplar-based texture synthesis technique. The confidence value of each pixel and image isophotes are combined to determine the priority of filling. Ref.~\cite{khamele2015approach} presents an image inpainting technique to remove occluded pixels when the occlusion is small. More specifically, it combines feature extraction and fast weighted-principal component analysis~(FW-PCA) to restore the occluded images. More recently, a hybrid technique~\cite{vijayalakshmi2017recognizing} has been proposed where a PDE method and modified exemplar inpainting is utilized to remark the occluded face region. However, the occlusion type of face images studied in this work is not representative of the real scenario.

\subsubsection{Deep learning for blind inpainting} Generative models are known for the ability to synthesize or generate new samples from the same distribution of the training dataset. The core problem in generative models is to address density estimation by unsupervised learning, which can be carried out by explicit density estimation~\cite{oord2016pixel,kingma2013auto}, or implicit density estimation~\cite{goodfellow2014generative}. The most prominent generative models are originated from the variational autoencoder~(VAE)~\cite{kingma2013auto} and the generative adversarial network~(GAN)~\cite{goodfellow2014generative,goodfellow2016nips}. The traditional autoencoder is used to reconstruct data, and the VAE~\cite{kingma2013auto} applies probabilistic spin to the traditional autoencoder to allow generating a new image. Having assumed that training data is generated from the underlying latent representation, the VAE derives a lower bound on the data likelihood that is tractable and can be optimized. This method is a principled approach to generative models, and useful features can be extracted by inference of $q(z|x)$. However, generated images are blurrier and of low quality. Instead, the GAN~\cite{goodfellow2014generative} learns to generate a new image from training distribution using a minimax game between the generator network and the discriminator network. The discriminator wants to distinguish between real and fake images, while the generator wants to fool the discriminator by generating real-looking images. Ref.~\cite{radford2015unsupervised} uses convolutional architectures in the GAN and can generate realistic samples. However, the GAN is notorious for unstable training characteristics. Makhzani et al.~\cite{makhzani2015adversarial} propose an adversarial autoencoder~(AAE) to use the GAN framework as a variational inference in a probabilistic autoencoder to ensure that generating from any part of prior space results in meaningful samples.

There are GAN variants for all kinds of applications. We focus on methods that are relevant to face image editing and image inpainting. One is a blind-inpainting work~\cite{xie2012image} that combines sparse coding~\cite{elad2006image} and deep neural networks to tackle image denoising and inpainting. In particular, a stacked sparse denoising autoencoder is trained to learn the mapping function from generated corrupted noisy overlapping image patches to the original noise-free ones. The network is regularized by a sparsity-inducing term to avoid over-fitting. This method does not need prior information about the missing region and provides solutions to complex pattern removal like the superimposed text from an image. Context Encoders~\cite{pathak2016context} combine the encoder-decoder architecture with context information of the missing part by regarding inpainting as a context-based pixel prediction problem. Specifically, the encoder architecture is trained on the input images with missing parts to obtain a latent representation while the decoder architecture learns to recover the lost information by using the latent representation. Pixel-wise reconstruction loss and adversarial loss are jointly used to supervise the context encoders to learn semantic inpainting results. Several variants of context encoders~\cite{pathak2016context} are proposed: some extend it by defining global and local discriminators~\cite{li2017generative,iizuka2017globally} and some take the result of context encoders as the input and apply joint optimization of image content and texture constraints to avoid visible artifacts around the border of the hole~\cite{yang2017high}. Ref.~\cite{ulyanov2018deep} relies on the power of the generative network to complete the corrupted image without requiring an external training dataset. A partial convolution based network~\cite{liu2018image} is proposed to only consider valid pixels and apply a mechanism that can automatically generate an updated mask, resulting in robustness to image inpainting for irregular holes. Information Maximizing Generative Adversarial Networks~(InfoGAN)~\cite{chen2016infogan} maximize the mutual information between latent variables and the observation in an unsupervised way. It decomposes the input noise vector into the source of incompressible noise $z$ and the latent code $c$ which will target the salient structured semantic features of data distribution. By manipulating latent codes, several visual concepts such as different hairstyles, presence or absence of eyeglasses are discovered. Occlusion-aware GAN~\cite{chen2017occlusion} is proposed to identify a corrupted image region with an associated corrupted region recovered using a GAN pre-trained on occlusion-free faces. Ref.~\cite{chen2018eyes} employs GAN for eyes-to-face synthesis with only eyes visible. Very recently, AttGAN\cite{he2019attgan}, a face image editing method, has imposed attribute classification constraints to the generated image so that the desired attributes are incorporated. Hairstyles and eyeglasses that may cause occlusion in a face image are treated as attributes which can be triggered to be present or absent in the generated image. ERGAN~(Eyeglasses removal generative adversarial network)~\cite{hu2019unsupervised} is proposed for eyeglasses removal in the wild in an unsupervised manner. It is capable of rendering a competitive removal quality in terms of realism and diversity. In paper~\cite{ge2020occluded}, ID-GAN~(identity-diversity generative adversarial network) combines a CNN-based recognizer and GAN-based recognition to inpaint realism and identity-preserving faces. The recognizer is treated as the third player to compete with the generator. 


\section{Performance evaluation}
In this section, we first evaluate the performance of occluded face detection on the MAFA dataset~\cite{ge2017detecting} of partially occluded faces. Next, we present the performance of face recognition under occlusion in terms of the identification rate and the verification accuracy on multiple benchmarks such as the AR~\cite{MaB1998the}, CAS-PEAL~\cite{gao2008cas}, and Extended Yale B~\cite{georghiades2001few} datasets. Then we describe the representative algorithms based on the proposed categories. In addition, we also categorize them in the face recognition pipeline according to which component they work on to tackle the occlusion. 

\subsection{Evaluation of Occluded Face Detection}

There are several datasets for face detection benchmarking, which are the FDDB~\cite{jain2010fddb}, PASCAL Face~\cite{yan2014face}, AFW~\cite{zhu2012face}, IJB-A~\cite{klare2015pushing}, Widerface~\cite{yang2016wider} and MAFA~\cite{ge2017detecting} datasets. MAFA is created for occluded face detection, involving 60 commonly used masks, such as simple masks, elaborate masks, and masks consisting of parts of the human body, which occur in daily life. Besides, it contains $35,806$ face annotations with a minimum size of $32\times 32$ pixels. Some examples of occluded face images are shown in Fig.~\ref{fig:occlusion_examples}. To the best of our knowledge, the MAFA dataset takes the occlusions as the main challenge in face detection, so it is relevant to evaluate the capacity of occluded face detection methods. The performances of representative algorithms on the MAFA dataset are summarized in Table~\ref{tab:ofdTab}~(derived from paper~\cite{chen2017masquer}). The precision on the MAFA testing set with the IoU threshold value $0.5$ is shown. Only a few methods report the results below.


\begin{table}[!htbp]
\renewcommand\arraystretch{1.3}
\centering
\caption{Evaluation Summary of Different Occluded Face Detection Algorithms on MAFA. The setting `masked' only counts the faces annotated by the MAFA dataset and `w/o Ignored' counts all detected faces, including the ones that are missing annotation. }\label{tab:ofdTab}
\begin{tabularx}{0.49\textwidth}{p{2cm}Xp{1.5cm}p{1.6cm}}
\toprule
 Approach  &Publication  & Precision `masked' & Precision `w/o Ignored' \\ \midrule
Occlusion unaware&  \cite{zhang2016joint} &    NA & $60.8\%$ \\
\midrule
\multirow{2}{*}{Occlusion aware}& \cite{ge2017detecting} &  NA & $76.4\%$\\
& \cite{wang2017face}  & $76.5\%$ & $88.3\%$ \\
\midrule
Occlusion segmentation & \cite{chen2017masquer} &$83.5\%$ & $91.9\%$ \\ \bottomrule
\end{tabularx}
\end{table}

\subsection{Existing OFR benchmark datasets}
There are numerous standard datasets for general face recognition, but they are not appropriate for OFR~(occluded face recognition) because occluded faces are barely present in the datasets. Alternatively, researchers make the most of the general datasets to generate synthetic occluded datasets by incorporating synthetic occlusions, occluding rectangles, etc. Five categories regarding OFR testing scenarios are illustrated in Fig.~\ref{fig:OFR_problems}. In this way, synthetic occluded datasets can meet the requirements of OFR, where an occlusion-free gallery is queried using occluded faces. 

\begin{table}[!htbp]
\renewcommand\arraystretch{1.4}
\centering
\caption{Summary of existing benchmark datasets for occluded face recognition. `\# Sub' is short for the number of subjects. `\# Ims' means the number of images. `Real Occ.' says whether real occlusions are included in the original dataset. `Syn. Occ.' is short for synthesized occlusions imposed on the original dataset to obtain `D-occ'.}\label{tab:dataset}
\begin{tabular}{l c c l l } 
\toprule
Dataset  & \# Sub & \# Ims &Real Occ. &Syn. Occ. \\ \midrule
AR~\cite{MaB1998the} & $126$ &$4,000$ &Yes &NA\\
Extend YaleB~\cite{georghiades2001few} &$38$&$161,289$ &No & Unrelated Ims\\
ORL~\cite{samaria1994parameterisation}&$40$&$400$&No&Rectangle \\
FERET~\cite{phillips2000feret}&$1500$&$13,000$&No&Rectangle\\
LFW~\cite{LFWTech}&$5749$&$13,000$&No&Partial Faces \\ 
CAS-PEAL~\cite{gao2008cas}&$1040$&$9,594$&Yes& NA\\ 
CMU-PIE~\cite{gross2010multi}&$68$&$41,368$&No&Rectangle\\
PubFig~\cite{kumar2009attribute}&$200$&$ 58,797$&No&Partial Faces\\
NIR-Distance~\cite{he2016multiscale}&$276$&$4300$&No&Partial Faces\\
NIR-Mobile~\cite{zhang2018deep}&$374$&$11,286$&No&Partial Faces\\
FRGC~\cite{frgc}&$276$&$943$&No&Partial Faces\\
FRGC-v2~\cite{phillips2005overview}&$466$&$4,007$&No&Partial Faces\\
\bottomrule
\end{tabular}
\end{table}

We make it a rule to state the `D' as the original dataset, and `D-occ' as the occluded dataset originated from `D'. If there is real occlusion included in the `D', then it is directly used for establishment. Otherwise, synthesized occlusions are imposed to form `D-occ' for establishment. Table~\ref{tab:dataset} summarizes each dataset by the number of subjects, the number of images, whether or not real occlusions are included, and if not, what kind of synthesized occlusions are used to obtain the `D-occ' dataset. Here AR and Extended Yale B, the most widely used face datasets, are discussed.

\textbf{AR} face database is one of the very few datasets that contain real occlusions~(see Fig.~\ref{fig:OFR_problems} first column). It consists of over $4000$ faces of $126$ individuals: $70$ men and $56$ women, taken in two sessions with a two week interval. There are $13$ images per individual in every session, and these images differ in terms of facial expression, illumination, and partial occlusion, getting sunglasses and scarves involved. Index $8$ and $11$ of each session indicates the person wearing sunglasses or a scarf,  respectively. Index $9-10$ and $11-12$ combine the sunglasses or the scarf with illuminations, respectively. 

\textbf{Extended Yale B} face database contains $161,289$ face images from $38$ persons, each captured under nine poses and 64 different illuminations without occlusion. It is widely used to evaluate the efficacy of the algorithms under synthesized occlusions. Occluding unrelated images are randomly superimposed on the occlusion-free faces~(see Fig.~\ref{fig:OFR_problems} last column). Typically, gallery images are occlusion-free faces, and test images are randomly occluded with unrelated images such as a baboon, or a square image. Sometimes, occluded faces are occluded with, for example white or black rectangles to constitute a `D-occ' dataset~(see Fig.~\ref{fig:OFR_problems} fourth column).

\subsection{Evaluation of Occluded Face Recognition}
In this part, we demonstrate the results of the most representative methods based on the proposed categories. In addition, we also categorize methods based on the face recognition pipeline and show evaluation results from this aspect.


\subsubsection{Evaluation based on the proposed categories} 
We summarize the identification performances of representative algorithms on the AR database in Table~\ref{tab:ARidenrate}. To make sure these methods can be compared, we group the experimental settings and introduce the corresponding abbreviation for simplicity as follows:
\begin{itemize}
    \item[(1)] Face images for training and testing belong to the same individual and without overlapping. We use `S-TR-$X$ips' to abbreviate the situation when the training set contains $X$ unoccluded images per person. Specifically, `S-TR-SSPP' single sample per person in the training set remains a challenging task. Usually, the gallery set keeps the same training set if there are no additional announcements. We add asterisks to mark the gallery, only taking one neutral face for enrollment.

    \item[(2)] Face images for training and testing belong to the same individual, without overlapping. The training set consists of occluded faces for training, denoting as `S-TR-Occ$X$', with $X$ as the number of images. Typically, one neutral face image per individual is enrolled.
    
    \item[(3)] Testing subjects are a subset of training subjects denoting as `Ex-TR-SSPP'. Images of extra persons are included in the training set. Apart from that, the same subjects are used for training and testing. The training set also consists of a single sample per person.
    
    \item[(4)] There is no overlapping between subjects for training and testing, which is represented by `D-TR'. We use the setting `D-TR-DL' to indicate if there are large-scale general data for training with deep learning techniques. 
    
    \item[(5)] Images of subjects are randomly selected for training, and the remaining ones are for testing, what we called `S-TR-Rand'.
\end{itemize}

The identification performance of representative algorithms on extended Yale B as well as other benchmarks is summarized in Table~\ref{tab:otheridenrate}. An evaluation summary of different categories of representative algorithms as regards verification rates is exhibited in Table~\ref{tab:allverirate}.


As results on the AR dataset in Table~\ref{tab:ARidenrate} show, these experimental setups were slightly different from paper to paper, bringing a struggle to interpret these results at the first sight. This is because the AR dataset does not provide standard protocols for use, which are essential to compare methods in a fair ground. Despite the absence of the standard protocols, it is still possible to get some useful observations and findings. First, most methods treat session1 of AR as the target and report the results on sunglasses and scarf individually. Second, there are some popular protocols gradually formed among these methods, which can be treated as a relatively fair ground for comparison. These are (i)~S-TR-7IPS, (ii)~S-TR-8IPS, and (iii)~S-TR-SSPP. Specifically, S-TR-SSPP in particular has become a challenge and defines an unambiguous protocol to follow. Last but not least, thanks to the presence of deep learning methods, there is a trend to rely on general massive data for training rather than splitting the AR dataset to form the training set. There is still some room to improve the OFR performance, especially when we take the SSPP~(single sample per person) protocol into consideration.

Results on Extended Yale B and other benchmark datasets are shown in Table~\ref{tab:otheridenrate} and Table~\ref{tab:allverirate}. Even though the original occlusion-free datasets such as Extended Yale B and ORL are well addressed by current face recognition algorithms, the `D-occ'~(occluded version) originated from these datasets still remains a challenge. It is not surprising that algorithms suffer severe degradation when $90\%$ of the face parts are occluded by a rectangle or unrelated images. Moreover, these testing scenarios are not similar to what we would find in a realistic scenario. 
Apart from an occluding rectangle and unrelated images, some methods intend to solve partial face issues by using arbitrary patches cropped from a face image. Since partial faces are arbitrary patches, it is hard to be sure algorithms are on a level playing field. For example, partial faces with eye regions are supposed to contain more discriminative features compared to those with no eyes involved.

\subsubsection{Evaluation based on the face recognition pipeline for categorization}
Generally, occlusion robust feature extraction~(ORFE) methods deal with occlusion using feature extraction. However, they sometimes treat occluded face recognition as a semantic correspondence problem and address the occlusion challenge in the comparison component of the face recognition pipeline. Similarly, occlusion aware face recognition~(OAFR) usually solves the occluded face recognition using comparison. Nevertheless, some occlusion-aware techniques use an occlusion-robust dictionary to represent an occluded face, which lies in the feature extraction component. Therefore, we categorize methods discussed so far based on a general face recognition pipeline to offer a fresh perspective~(see Table~\ref{tab:components}). Each component is elaborated as follows:

\begin{itemize}
    \item Data Augmentation or Data Recovery: augmentation based techniques cope with the occlusion challenge by augmenting training data to include possible real-life occlusion. With the use of deep learning techniques, occlusion-robust features can be extracted. Recovery based methods use local or global knowledge to explicitly fill the occlusion area or de-occlude the face in an implicit way, for example using an encoder-decoder structure. 
    
    \item Feature extraction intends to make the most of the locality characteristic of occlusion using patch-based engineered feature extraction or applies machine learning or deep learning to obtain features. Specifically, statistical learning or sparse representation are two commonly used techniques in this approach.
    
    \item Feature comparison: features obtained for an entire face are of fixed length, while features for occluded face images are determined by the occlusion area and are therefore of varied length. The comparison phase is used to find out the semantic correspondence between fixed-length and varied features because it is meaningless to match different semantic parts.
    
    \item Fusion strategy relies on the assumption that occlusions have less effect on identity recognition compared with other visible facial areas. Occlusion challenges can be met by fusing the decision of multiple classifiers or using the majority voting strategy to minimize the effect of features extracted from the occlusion.
\end{itemize}

\section{Discussion}
In this section, future dataset challenges and research challenges are discussed. In many cases, a new research challenge means specific dataset requirements. On the other hand, datasets also reflect underlying challenges that need to be solved in real life.

\subsection{Future Dataset Challenges}
There are three major issues in the datasets referred to: \textbf{dataset scale}, \textbf{occlusion diversity}, and \textbf{standard protocol}. The datasets for occluded face recognition are of a small scale. AR is one of the very few that contain real occlusions and images of only $126$ individuals are included. As for occlusion diversity, sunglasses and scarf occlusions are frequently considered. However, occlusions in real life are a lot more diverse than that. Regarding synthesized occlusions, occluding rectangles and unrelated images are commonly applied. In the literature, ‘Baboon’ is a typically used unrelated image to generate synthetic occluded faces. However, these synthetically occluded faces like those using ‘Baboon’, or a black/white rectangle are not representative of real-life occluded faces. As regards a standard protocol, the results of representative approaches obtained from AR and other benchmark datasets cannot be compared objectively due to a wide diversity of occluded faces. Therefore, future research will have to overcome three weaknesses in the current datasets. Future benchmark datasets will contain more individuals, various and sufficient occlusions, and well-defined evaluation protocols. 

Apart from specific requirements for benchmark datasets, collecting a massive occluded face dataset for training is also inevitable as we plunge into deep learning techniques. However, the main source of face images is usually the web, where labelled faces with occlusions are scarce. How to label occluded faces efficiently is an open issue that needs to be resolved because some severely occluded faces are difficult or impossible to be recognized by humans. From this aspect, occluded face recognition remains a challenge in the future. Given the combination of the pose variations, low resolution etc., recognizing unconstrained faces is far from being solved.

For the development of occluded face recognition techniques, new large-scale datasets are needed that respond to the occlusion challenges. Recently, IJB-C, a large-scale dataset, has been made publicly available. It not only contains many natural occlusions but also provides annotated occlusion locations. The public availability of IJB-C may usher in a new development of occluded face recognition. In the future, we expect to see the standard evaluation of fully efficient occluded face recognition algorithms on the AR dataset as well as on newly developed real-life and large-scale IJB-C face datasets.

\subsection{Future Research Challenges}
In the future, unconstrained occluded face recognition will become a challenging problem that needs to be addressed. Unlike the occluded faces we are handling, unconstrained occluded faces should conform to the future benchmark datasets, not limited by predefined occlusions. Currently used datasets such as AR and Extended Yale B are both problematic since they are aligned and captured in a constrained environment. Apart from unconstrained occluded face recognition, it remains an open issue to resolve face recognition burdened with unconstrained occlusions and other challenges such as large pose variations, age gap, low resolution, etc. It is worth mentioning that unconstrained face recognition usually does not take the occlusion challenge as the major challenge. Instead, unconstrained face recognition mainly takes pose and expression challenges into consideration. Face recognition will be free from constraints if real-life occlusion plays an equally important role among these challenges. 

Since the existing occluded face recognition approaches cope with occlusion challenges from distinct perspectives, we expect the individual improvements in occlusion robust feature extraction, occlusion aware face recognition, and occlusion recovery based face recognition. The majority of methods in the literature so far follow the line of traditional machine learning techniques. Methods for occluded face recognition in the future will make use of deep learning techniques, especially CNN based deep networks. Based on the method categories in Fig.~\ref{fig:methods_category}, we suggest a number of ways to develop a face recognition system that is better able to handle occlusions as follows.

\textbf{Novel data augmentation techniques} are needed that could help to learn more discriminative feature representations robust to occlusions if a CNN deep network is trained on the augmented dataset. Current solutions for data augmentation mainly rely on large-scale training data from the web, and use occlusion templates to generate synthesized occluded faces. However, manually designed occlusion templates heavily rely on accurate facial landmarking, and an acceptable occluded face can only be achieved if the occlusion template is properly aligned with the facial landmarks. For example, eyeglasses should be scaled and aligned to the eye centers. One potential solution is to take advantage of GANs to generate natural occluded faces by transferring the occlusion type from a real occluded face to an occlusion-free face.

\textbf{More effective occlusion recovery models} need to be devised to make the most of state-of-the-art unified face recognition systems. Current occlusion recovery solutions adopt an encoder-decoder architecture for recovery and are designed for visually pleasing results rather than accurate reconstruction. One potential way to resolve this issue is to combine the occlusion recovery task and identification or verification task in a multi-task manner. The adversarial loss can be incorporated to ensure that the recovered face looks natural.

\textbf{New occlusion detection techniques} need to be designed to take advantage of massive public datasets. Better occlusion detection not only makes it possible to generate occlusion templates automatically but also provides an informative preprocessing for face recognition. Most methods reviewed address the occlusion detection problem by partitioning the face into patches using a binary classifier, resulting in a rough occlusion area. Combining occlusion detection and occluded face recognition in a unified framework seems a promising way, leading to an automatic face recognition system. 

\section{Conclusion}
In this paper, we present a thorough survey of face recognition techniques under occlusion and systematically categorize methods up to now into 1) occlusion robust feature extraction, 2) occlusion aware face recognition, and 3) occlusion recovery based face recognition. Newly published and innovative papers, especially recent deep learning techniques for occluded face recognition, have been discussed. Furthermore, we report comparative performance evaluations in terms of occluded face detection and face recognition. In the end, we discuss future challenges in terms of dataset and research~(including potential solutions) that move the field forward.



\begin{table*}[htbp]
\renewcommand\arraystretch{0.9}
\caption{Evaluation Summary of Representative Algorithms on AR dataset regarding identification rates. Three categories are: 1)~\textbf{ORFE}=Occlusion Robust Feature Extraction, 2)~\textbf{OAFR}=Occlusion Aware Face Recognition, and 3)~\textbf{ORecFR}=Occlusion recovery FR. The detail of `Experiment settings' abbreviations is shown. In column `Occlusions,' the asterisk means specific occlusion under illumination. In column `Identification Rates,' TR means these methods train on session one.}\label{tab:ARidenrate}
\centering
\begin{tabularx}{\textwidth}{lXlXXccc}
\toprule
\multirow{2}{*}{Category}  & \multirow{2}{*} {Publication} & \multirow{2}{*}{Occlusions}&\multirow{2}{*}{\begin{tabular}[c]{@{}l@{}}Experiment\\Settings\end{tabular}} &\multirow{2}{*}{\begin{tabular}[c]{@{}c@{}}\# of Train\\Subjects\end{tabular}}&\multirow{2}{*}{\begin{tabular}[c]{@{}c@{}}\# of Test\\Subjects\end{tabular}}& \multicolumn{2}{c}{Identification Rates (\%)} \\
\cline{7-8}
  &&&&&&Session1& Session2\\
\midrule
\multirow{4}{*}{\raggedright \begin{tabular}[c]{@{}l@{}l@{}}\textbf{ORFE:}\\Patch-engineered\\Features\end{tabular} } 
 &\multirow{2}{*}{\cite{zhang2007local}} & Sunglasses & \multirow{2}{*}{S-TR-$8$IPS}&\multirow{2}{*}{50} & \multirow{2}{*}{50} &84.0&80.0\\
 &&Scarf & & &  & 100.0& 96.0\\
&\multirow{2}{*}{\cite{cheheb2017random}}& Sunglasses &  \multirow{2}{*}{S-TR-SSPP}&\multirow{2}{*}{100} & \multirow{2}{*}{100} & 89.0&98.0\\
&&Scarf&&&&73.0&92.0\\
\midrule
\multirow{30}{*}{\raggedright \begin{tabular}[c]{@{}l@{}l@{}}\textbf{ORFE:}\\Learning-based\\Features\end{tabular}} 
&\multirow{2}{*}{\cite{martinez2002recognizing}} & Sunglasses &  \multirow{2}{*}{S-TR-SSPP}&\multirow{2}{*}{50} &\multirow{2}{*}{50} & 52.0&NA\\
&&Scarf&&&&48.0&NA\\
&\multirow{2}{*}{\cite{hotta2008robust}} &Sunglasses& \multirow{2}{*}{S-TR-$8$IPS}&\multirow{2}{*}{40} &\multirow{2}{*}{40} & 80.0&NA\\
&&Scarf&&&&70.8&NA\\
\cline{2-7}
&\multirow{2}{*}{\cite{tan2009face}}&Sunglasses& \multirow{2}{*}{D-TR}&\multirow{2}{*}{50}&\multirow{2}{*}{50}&98.0&NA\\
&&Scarf&&&&90.0&NA\\
&\multirow{2}{*}{\cite{wright2009robust}}&Sunglasses& \multirow{2}{*}{S-TR-8IPS}& \multirow{2}{*}{100}&\multirow{2}{*}{100}&97.5&NA\\
&&Scarf&&&&93.5&NA\\
&\multirow{2}{*}{\cite{seo2011robust}}&Sunglasses\&Scarf& \multirow{2}{*}{S-TR-3IPS}&\multirow{2}{*}{100}&\multirow{2}{*}{100}&89.0&NA\\
&&Black block&&&& 94.0 &NA\\
&\multirow{2}{*}{\cite{yang2013gabor}}&Sunglasses& \multirow{2}{*}{S-TR-7IPS}&\multirow{2}{*}{100} &\multirow{2}{*}{100}& 92.3&51.7\\
&&Scarf&&&&95.0&84.3\\
&\multirow{3}{*}{\cite{ou2014robust}}&Sunglasses$^\star$& \multirow{2}{*}{S-TR-Occ1}&\multirow{3}{*}{100}&\multirow{3}{*}{100}&\multicolumn{2}{c}{93.0}\\
&&Scarf$^\star$&&&&\multicolumn{2}{c}{92.7}\\
&&Sunglasses\&Scarf$^\star$&&&&\multicolumn{2}{c}{92.6}\\
&\multirow{2}{*}{\cite{seo2014robust}}&Sunglasses$^\star$& \multirow{2}{*}{S-TR-3IPS}&\multirow{2}{*}{100}&\multirow{2}{*}{100}&95.7& NA\\
&&Scarf$^\star$&&&&96.3&NA\\
&\multirow{2}{*}{\cite{mclaughlin2017largest}}&Sunglasses$\star$& \multirow{2}{*}{S-TR-Occ3}&\multirow{2}{*}{100}&\multirow{2}{*}{100}&TR&98.0\\
&&Scarf$^\star$&&&&TR&97.0\\
&\multirow{3}{*}{\cite{gao2017learning}}&Sunglasses& \multirow{2}{*}{S-TR-SSPP}&\multirow{3}{*}{100}&\multirow{3}{*}{100}&TR& 92.0\\
&&Scarf&&&&TR& 91.0\\
&&Sunglasses\&Scarf&&&&TR& 82.5\\
&\multirow{2}{*}{\cite{li2017ubiquitous}}&Sunglasses$^\star$& \multirow{2}{*}{S-TR-SSPP}&\multirow{2}{*}{80}&\multirow{2}{*}{80}& 88.0&56.0\\
&&Scarf$^\star$&&&&69.0&44.0\\
&\multirow{2}{*}{\cite{yu2017discriminative}}&Sunglasses$^\star$& \multirow{2}{*}{EX-TR-SSPP}&\multirow{2}{*}{80} & \multirow{2}{*}{80}& 95.8& 78.3\\
&&Scarf$^\star$&&&&90.0&77.9\\
&\cite{yang2017joint}&Sunglasses\&Scarf&D-TR-DL&Webface+20 & 80 &100.0&96.3\\
&\cite{wu2018occluded}&Sunglasses\&Scarf&S-TR-SSPP&100&100&\multicolumn{2}{c}{75.0}\\
&\multirow{2}{*}{\cite{cen2019dictionary}}&Sunglasses& \multirow{2}{*}{S-TR-7IPS}&\multirow{2}{*}{50} & \multirow{2}{*}{50}& 94.7& 85.3\\
&&Scarf&&&&99.3&98.7\\
&\multirow{2}{*}{\cite{ge2020occluded}}&Sunglasses&\multirow{2}{*}{S-TR-3IPS}&\multirow{2}{*}{80}&\multirow{2}{*}{80}&99.0&NA\\
&&Scarf&&&&87.6&NA\\

\midrule
\multirow{8}{*}{\raggedright \begin{tabular}[c]{@{}l@{}l@{}}\textbf{OAFR:}\\Occlusion Detection\\Face Recognition\end{tabular}}
&\multirow{2}{*}{\cite{oh2008occlusion}}&Sunglasses& \multirow{2}{*}{EX-TR-SSPP}&\multirow{2}{*}{100}&\multirow{2}{*}{100}&NA&49.0\\
&&Scarf&&&&NA&55.0\\
&\multirow{2}{*}{\cite{min2011improving}}&Sunglasses$^\star$& \multirow{2}{*}{S-TR-3IPS$^\star$}&\multirow{2}{*}{80}&\multirow{2}{*}{80}& TR&54.2\\
&&Scarf$^\star$&&&&TR&81.3\\
&\multirow{2}{*}{\cite{andres2014face}}&Sunglasses$^\star$& \multirow{2}{*}{S-TR-8IPS-Occ1}&\multirow{2}{*}{60}&\multirow{2}{*}{60}&97.5&NA\\
&&Scarf$^\star$&&&&99.2&NA\\
&\multirow{2}{*}{\cite{ou2018robust}}&Sunglasses& \multirow{2}{*}{S-TR-RAND}&\multirow{2}{*}{100}&\multirow{2}{*}{100}&\multicolumn{2}{c}{95.2}\\
&&Scarf&&&&\multicolumn{2}{c}{94.2}\\
&\multirow{2}{*}{\cite{song2019occlusion}}&Sunglasses&\multirow{2}{*}{D-TR-DL}&\multirow{2}{*}{Webface}&\multirow{2}{*}{100}&99.7&NA\\
&&Scarf&&&&100.0&NA\\
\midrule
\multirow{8}{*}{\raggedright \begin{tabular}[c]{@{}l@{}}\textbf{OAFR:}\\Partial Face Recognition\end{tabular}}
&\multirow{3}{*}{\cite{hu2013robust}}&Sunglasses$^\star$& \multirow{2}{*}{S-TR-$7$IPS}&\multirow{3}{*}{100}&\multirow{3}{*}{100}&94.3&NA\\
&&Scarf$^\star$&&&&98.0&NA\\
&&Sunglasses\&Scarf$^\star$&&&&96.2&NA\\
&\multirow{3}{*}{\cite{weng2013robust}}&Sunglasses$^\star$& \multirow{2}{*}{S-TR-$14$IPS}&\multirow{3}{*}{100}&\multirow{3}{*}{100}&98.0&NA\\
&&Scarf$^\star$&&&&97.0&NA\\
&&Sunglasses\&Scarf$^\star$&&&&97.5&NA\\
&\multirow{2}{*}{\cite{weng2016robust}}&Sunglasses$^\star$& \multirow{2}{*}{S-TR-$7$IPS}&\multirow{2}{*}{100}&\multirow{2}{*}{100}&100.0&92.0\\
&&Scarf$^\star$&&&&100.0&95.3\\
\midrule
\midrule
\multirow{12}{*}{\raggedright \begin{tabular}[c]{@{}l@{}l@{}}\textbf{ORecFR:}\\Occlusion Recovery\\Face Recognition\end{tabular}}
&~\cite{jia2008face}&Sunglasses\&Scarf$^\star$&S-TR-13IPS&100&100&TR&90.6\\
&\multirow{2}{*}{\cite{zhou2009face}}&Sunglasses& \multirow{2}{*}{S-TR-$8$IPS}&\multirow{2}{*}{100}&\multirow{2}{*}{100}&100.0&NA\\
&&Scarf&&&&97.0&NA\\
&\cite{deng2011graph}&Sunglasses$^\star$&\multirow{2}{*}{S-TR-$7$IPS}&\multirow{2}{*}{121}&\multirow{2}{*}{121}&76.6&NA\\
&&Scarf$^\star$&&&&60.9&NA\\
&\cite{li2013reconstruction}&Sunglasses$^\star$& \multirow{2}{*}{S-TR-$8$IPS}&100&100&97.5&NA\\
&\multirow{2}{*}{\cite{luan2014extracting}}&Sunglasses& \multirow{2}{*}{S-TR-$8$IPS}&\multirow{2}{*}{100}&\multirow{2}{*}{100}&94.5&NA\\
&&Scarf&&&&95.0&NA\\
&\multirow{2}{*}{\cite{zhao2015modular}}&Sunglasses&\multirow{2}{*}{S-TR-$3$IPS-Occ3}&\multirow{2}{*}{120}&\multirow{2}{*}{120}&NA&68.5\\
&&Scarf&&&&NA&70.7\\

&\multirow{2}{*}{\cite{iliadis2017robust}}&Sunglasses& \multirow{2}{*}{S-TR-$2$IPS}&\multirow{2}{*}{100}&\multirow{2}{*}{100}&\multicolumn{2}{c}{89.8}\\
&&Scarf&&&&\multicolumn{2}{c}{78.8}\\

&\multirow{2}{*}{\cite{wang2017varying}}&Sunglasses&\multirow{2}{*}{S-TR-$4$IPS}&\multirow{2}{*}{120}&\multirow{2}{*}{120}&99.2&99.7\\
&&Scarf&&&&87.5&$83.6$\\
\bottomrule
\end{tabularx}
\end{table*}

\begin{table*}[!htbp]
\renewcommand\arraystretch{0.95}
\caption{Evaluation Summary of Representative Algorithms on the other Benchmarks regarding identification rates. Three categories: 1)~\textbf{ORFE}=Occlusion Robust Feature Extraction, 2)~\textbf{OAFR}=Occlusion Aware Face Recognition, and 3)~\textbf{ORecFR}=Occlusion recovery FR. In the `Occlusion Arguments' column, details including the size of occlusion, the occlusion ratio to the face image and portion of noise are listed. In `\# of Train Subjects,' NA represents not available from the paper. The Rank 1 identification rates are shown. Some results are roughly estimated from the figure in the original papers. Since not all methods are of the same experimental settings, these methods cannot be compared properly. }\label{tab:otheridenrate}
\centering
\begin{tabularx}{\textwidth}{llXllcccX}
\toprule
Category & Publication & Occlusions&\begin{tabular}[c]{@{}l@{}}Occlusion Arguments \end{tabular}& Benchmark&\begin{tabular}[c]{@{}c@{}}\# of Train\\Subjects\end{tabular}& \begin{tabular}[c]{@{}c@{}}\# of Test\\Subjects\end{tabular}& \begin{tabular}[c]{@{}c@{}}Identification \\Rates(\%)\end{tabular}\\
\midrule
\multirow{20}{*}{\raggedright\textbf{ORFE}}&\cite{hotta2008robust}&Black or White Rectangle &occlusion size:$50\times50$&ORL&40&40&70.0\\
&\cite{tan2009face}&Black Rectangle &occlusion size:$50\times 50$&FERET&240&960&$78.5$\\
&\cite{wright2009robust}&Unrelated image &occlusion ratio:0.5&Extended Yale B&38&38&$65.3$\\
&\cite{yang2013gabor}&Unrelated image &occlusion ratio:0.5&Extended Yale B&38&38&$87.4$\\
&\cite{ou2014robust}&Unrelated image &occlusion ratio:0.8&Extended Yale B&38&38&$70.0$\\
&\cite{mclaughlin2017largest}&Unrelated image &occlusion ratio:0.8&Extended Yale B&38&38&$53.0$\\
&\multirow{2}{*}{\cite{yang2017joint}}&\multirow{2}{*}{Wild condition}&NA&\multirow{2}{*}{LFW}& 108 &50&$86.0$\\
&&&NA&&851&124&$65.3$\\
&\multirow{3}{*}{\cite{yu2017discriminative}}&Glasses\&Sunglasses&NA&\multirow{2}{*}{CAS-PEAL}&\multirow{2}{*}{350}&\multirow{2}{*}{350}&$96.6$\\
&&Hat&NA&&&&$90.3$\\
&&Unrelated image &occlusion ratio:0.5&Extended Yale B&30&30&78.4\\
&\multirow{3}{*}{\cite{gao2017learning}}&Unrelated image &occlusion ratio:0.6&\multirow{3}{*}{Extended Yale B}&\multirow{3}{*}{38}&\multirow{3}{*}{38}&$96.0$\\
&&Random pixels Rectangle &occlusion ratio:0.6&&&&81.0\\
&&Mixture noise&occlusion ratio:0.6&&&&$25.0$\\
&\cite{li2017ubiquitous}&Unrelated image &occlusion ratio:0.3&Extended Yale B&30&30&$77.6$\\

&\multirow{4}{*}{\cite{ou2018robust}}&Pepper noise&portion:40\%&\multirow{2}{*}{Extended Yale B}&\multirow{2}{*}{38}&\multirow{2}{*}{38}&$81.3$\\
&&White Rectangle&NA&&&&$82.9$\\
&&Salt\&Pepper noise &portion:40\%&\multirow{2}{*}{CMU-PIE}&\multirow{2}{*}{68}&\multirow{2}{*}{68}&$98.5$\\
&&White Rectangle&occlusion size:$10\times10$&&&&$98.8$\\

&\multirow{3}{*}{\cite{du2019nuclear}}&Black Rectangle &occlusion ratio:0.3&\multirow{3}{*}{Extended Yale B}&\multirow{3}{*}{38}&\multirow{3}{*}{38}&$99.2$\\
&&Black Rectangle&occlusion ratio:0.4&&&&$97.6$\\
&&Black Rectangle&occlusion ratio:0.5&&&&$96.1$\\

&\multirow{3}{*}{\cite{fu2017efficient}}&Unrelated image &occlusion ratio:0.5&\multirow{2}{*}{Extended Yale B}&\multirow{2}{*}{38}&\multirow{2}{*}{38}&$96.9$\\
&& Random pixels Rectangle&portion:70\%&&&&$98.9$\\
&&Black Rectangle&occlusion ratio:0.5&CMU-PIE&68&68&$93.9$\\

\midrule
\midrule
\multirow{10}{*}{\raggedright \textbf{OAFR} } &\cite{hu2013robust} &Arbitrary Patches&NA&Partial-LFW&158&158&$34.8$\\
&\multirow{2}{*}{\cite{weng2013robust}} &Arbitrary Patches&NA&Partial-LFW&158&158&$50.7$\\
&&Unrelated image &occlusion ratio:0.5&Extended Yale B&32&32&$30.2$\\
&\multirow{2}{*}{\cite{weng2016robust}}&Unrelated image &occlusion ratio:0.5&Extended Yale B&32&32&$56.7$\\
&&Arbitrary Patches&NA&Partial-PubFig&60&140&$42.9$\\
&\multirow{3}{*}{\cite{he2018dynamic}}&Arbitrary Patches&NA&Partial-LFW&NA&1000&$27.3$\\
&&Arbitrary Patches&NA&NIR-Distance&NA&276&$94.9$\\
&&Arbitrary Patches&NA&Partial-YTF&NA&200&$61.0$\\
&\multirow{3}{*}{\cite{he2019dynamic}}&Arbitrary Patches&NA&Partial-LFW&NA&1000&$32.4$\\
&&Arbitrary Patches&NA&NIR-Distance&NA&127&$92.8$\\
&&Arbitrary Patches&NA&NIR-Mobile&NA&178&$93.8$\\
\midrule
\midrule
\multirow{8}{*}{\raggedright \textbf{ORecFR} } &\cite{luan2014extracting}&Unrelated image &occlusion ratio:0.7&Extended Yale B&38&38&$62.3$\\
&\multirow{2}{*}{\cite{zhou2009face}}&Unrelated image &occlusion ratio:0.7&\multirow{2}{*}{Extended Yale B}&\multirow{2}{*}{38}&\multirow{2}{*}{38}&$88.5$\\
&&Multiple patches&occlusion ratio:0.8&&&&$96.0$\\
&\multirow{2}{*}{\cite{li2013reconstruction}}&Black or White Rectangle &occlusion ratio:0.57&\multirow{2}{*}{Extended Yale B}&\multirow{2}{*}{38}&\multirow{2}{*}{38}&$90.4$\\
&&Unrelated image &occlusion ratio:0.5&&&&$87.7$\\
&\cite{zhao2015modular}&Black rectangle on eyes&occlusion ratio:0.3&Extended Yale B&38&38&$98.6$\\
&\multirow{2}{*}{\cite{iliadis2017robust}}&Unrelated image &occlusion ratio:0.6&\multirow{2}{*}{Extended Yale B}&\multirow{2}{*}{38}&\multirow{2}{*}{38}&$95.8$ \\
&&Unrelated image &occlusion ratio:0.9&&&&$71.9$\\
\bottomrule
\end{tabularx}
\end{table*}

\begin{table*}[!htbp]
\caption{Evaluation Summary of Representative Algorithms regarding verification rates. Occlusion Robust Feature Extraction and Occlusion Aware Face Recognition report the verification rates by cut-off rules. In the `Benchmark' column, `Partial-x' means the new partial faces originate from the database named `x.' Some results are roughly estimated from the figure in the original papers. Since not all methods are of the same experimental settings, these methods cannot be compared properly.}\label{tab:allverirate}
\centering
\begin{tabularx}{0.68\textwidth}{lllcl}
\toprule
Publication & Occlusions&Benchmark& \# of Subjects in Gallery & Verification Rates\\
\midrule
\cite{savvides2006partial}&Eyeglasses& FRGC-v2 &466 & 90\%@FAR=0.1\%\\
\cite{hua2009robust}& Wild condition& LFW& 1680& 60\%@FAR=0.1\\
\cite{mclaughlin2017largest}&Wild condition&LFW&1680&61\%@FAR=0.1\\
\midrule
\midrule
\multirow{2}{*}{\cite{weng2016robust}}&Arbitrary Patches&Partial-LFW&1680&50\%@FAR=0.1\\
&Arbitrary Patches&Partial-PubFig&140&63\%@FAR=0.1\\
\cite{he2018dynamic}&Arbitrary Patches&Partial-LFW&1000&29.8\%@FAR=0.1\%\\
\cite{he2019dynamic}&Arbitrary Patches&Partial-LFW&1000&37.9\%@FAR=0.1\%\\
\bottomrule
\end{tabularx}
\end{table*}

\begin{table*}[!htbp]
\caption{Summary of Representative Algorithms based on components they work on during OFR. In the `data augmentation/recovery' category, data augmentation component means generating synthesized occluded faces while the data recovery component intends to eliminate the occluded facial part. The fusion strategy component consists of feature-level fusion as well as decision-level fusion.}\label{tab:components}
\renewcommand\arraystretch{1.3}
\centering
\begin{tabularx}{\textwidth}{|l|X|}
\hline
Pipeline Category & Publication\\
\hline
Data Augmentation/ Recovery&\cite{de2003framework,leonardis2000robust,fidler2006combining,jia2008face,lv2017data,park2005glasses,luan2014extracting,deng2011graph,li2013reconstruction,cheng2015robust,zhao2018robust,trigueros2018enhancing,osherov2017increasing,criminisi2004region,khamele2015approach,vijayalakshmi2017recognizing,chen2016infogan,he2019attgan,wang2017varying,hu2019unsupervised,ge2020occluded}\\
\hline
Feature Extraction&\cite{ahonen2004face,ahonen2006face,lowe1999object,dalal2005histograms,zou2007comparative,zhang2007local,kim2005effective,martinez2002recognizing,tan2005recognizing,hotta2008robust,seo2011robust,seo2014robust,wright2009robust,yang2013gabor,ou2014robust,yang2017joint,gao2017learning,yu2017discriminative,li2017ubiquitous,ou2018robust,liao2013partial,he2018dynamic,he2019dynamic,jia2008face,zhou2009face,iliadis2017robust,wan2017occlusion,song2019occlusion,cen2019dictionary,wen2016structured,fu2017efficient,zheng2020novel,du2019nuclear}\\
\hline
Feature Comparison&\cite{wiskott1997face,hua2009robust,tan2009face,mclaughlin2017largest,chen2011occluded,min2011improving,andres2014face,neo2010development,hu2013robust,weng2013robust,weng2016robust}\\
\hline
Fusion Strategy&\cite{pan2007part,savvides2006partial,cheheb2017random,oh2008occlusion,he2016multiscale,gutta2002investigation,tarres2005novel,zhao2015modular,zheng2020novel}\\
\hline
\end{tabularx}
\end{table*}

\begin{table*}[!htbp]
\caption{Summary of Representative Algorithms regarding application-oriented purpose from Fig.~\ref{fig:methods_category}. There is a category for occluded face detection with the abbreviation `\textbf{OFD}'. Three classes for occluded face recognition: 1)~\textbf{ORFE}=Occlusion Robust Feature Extraction, 2)~\textbf{OAFR}=Occlusion Aware Face Recognition, and 3)~\textbf{ORFR}=Occlusion recovery FR. To our knowledge, there are no research works on detecting partial face; thus, we mark it as `NA' (not available).}\label{apporiented}
\renewcommand\arraystretch{1.3}
\centering
\begin{tabularx}{\textwidth}{|c|l|X|}
\hline
Abbreviation&Application-oriented Purpose &Publication\\
\hline
\textbf{OFD}&Occluded Face Detection&\cite{chen2006modification,yang2015facial,jain2010fddb,yan2014face,jain2010fddb,zhu2012face,yang2018faceness,yang2015facial,yang2016wider,mahbub2016partial,samangouei2015attribute,zhang2015touch,opitz2016grid,ge2017detecting,wang2017face,chen2017masquer,wu2019hierarchical}\\
\hline
\multirow{3}{*}{}
\multirow{2}{*}{\textbf{ORFE}}& Patch based engineered features&\cite{ahonen2004face,ahonen2006face,lowe1999object,dalal2005histograms,zhang2005local,zou2007comparative,pan2007part,savvides2006partial,zhang2007local,kim2005effective,hua2009robust,cheheb2017random}\\
\cline{2-3}
 &Learning based features&\cite{leonardis2000robust,fidler2006combining,kim2005effective,lv2017data,martinez2002recognizing,tan2005recognizing,hotta2008robust,tan2009face,seo2011robust,seo2014robust,mclaughlin2017largest,wright2009robust,yang2013gabor,ou2014robust,yang2017joint,gao2017learning,yu2017discriminative,li2017ubiquitous,trigueros2018enhancing,osherov2017increasing,cen2019dictionary,wen2016structured,fu2017efficient,zheng2020novel,du2019nuclear}\\
\hline
\multirow{3}{*}{\textbf{OAFR}}
&Occlusion Detection& \cite{chen2017masquer,xia2015face}\\
\cline{2-3}
&Occlusion Discard Face Recognition&\cite{oh2008occlusion,chen2011occluded,min2011improving,andres2014face,neo2010development,ou2018robust,wan2017occlusion,song2019occlusion}\\
\cline{2-3}
&Partial Face Detection &NA\\
\cline{2-3}
&Partial Face Recognition&\cite{he2016multiscale,gutta2002investigation,liao2013partial,hu2013robust,weng2013robust,weng2016robust,he2018dynamic,he2019dynamic}\\
\hline
\multirow{2}{*}{\textbf{ORFR}}
&Occlusion Recovery&\cite{criminisi2004region,khamele2015approach,vijayalakshmi2017recognizing,xie2012image,elad2006image,pathak2016context,cheng2015robust,li2017generative,iizuka2017globally,yang2017high,ulyanov2018deep,liu2018image,chen2016infogan,he2019attgan,chen2018eyes,chen2017occlusion,hu2019unsupervised}\\
\cline{2-3}
&Occlusion Recovery Face Recognition&\cite{park2005glasses,de2003framework,leonardis2000robust,fidler2006combining,li2013reconstruction,luan2014extracting,jia2008face,zhou2009face,deng2011graph,zhao2015modular,iliadis2017robust,zhao2018robust,wang2017varying,ge2020occluded}\\
\hline
\end{tabularx}
\end{table*}

\appendix
A glossary of abbreviations and expansions for terminologies is illustrated in Table~\ref{tab:abbreviation}.

\begin{table*}[!ht]
\renewcommand\arraystretch{0.95}
\centering
\caption{A glossary of abbreviations and expansions for terminologies. }\label{tab:abbreviation}
\begin{tabularx}{\textwidth}{l l | l l } 
\toprule
Abbreviation  &  Expansion & Abbreviation  &  Expansion\\ \midrule
AAE & Adversarial AutoEncoder &LBP & Local Binary Patterns\\
AFC&Adaptive Feature Convolution&LDA & Linear Discriminant Analysis\\
AOFD & Adversarial Occlusion-aware Face Detection&LLE & Locally Linear Embedding\\
CSLBP&Center-Symmetric Local Binary Patterns&LSTM & Long Short Term Memory\\
DA& Denosing Autoencoder&MDSCNN&Multiscale Double Convolutional Neural Network\\
DCNNs & Deep Convolutional Neural Networks&MKD&Multi-Keypoint Descriptors\\
DDRC&Deep Dictionary Representation based Classification&NNAODL&Nuclear Norm based Adapted Occlusion Dictionary Learning\\
DMSC&Discriminative Multiscale Sparse Coding&NMF&Non-negative Matrix Factorization\\
DoG & Different of Gaussian filters &OAFR & Occlusion Aware Face Recognition\\
DPM & Deformable Part Models&OFR & Occluded Face Recognition\\
DSCNNs&Double Supervision Convolutional Neural Network &ORFE & Occlusion Robust Feature Extraction\\
EBGM & Elastic Bunch Graph Matching&ORecFR & Occlusion Recovery based Face Recognition\\
ECSLBP &Enhanced Center-Symmetric Local Binary Patterns&PCA & Principal Component Analysis\\
ELOC&Efficient Locality-constrained Occlusion Coding&PDSN&Pairwise Differential Siamese Network\\
ERBF&Ensemble of Radial Basis Function&R-CNN &Regions with Convolutional Neural Networks\\
ERGAN&Eyeglasses Removal Generative Adversarial Network& RCSLBP&Reinforced Centrosymmetric Local Binary Pattern \\
FAN & Face Attention Network&RDLRR&Robust Discriminative Low-Rank Representation\\
FANet & Feature Agglomeration Networks&RLA & Robust LSTM-Autoencoders\\
FCN&Fully Convolutional Network&RPSM&Robust Point Set Matching \\
FW-PCA&Fast Weighted-Principal Component Analysis&SIFT & Scale Invariant Feature Transform\\
GAN&Generative Adversarial Network&SOC&Structured Occlusion Coding\\
GRRC & Gabor based Robust Representation and Classification&SOM & Self-Organizing Maps\\
HOG & Histogram of Oriented Gradient&SRC & Sparse Representation Classifiers\\
ICA & Independent Component Analysis&SSD & Single-Shot multibox Detector\\
ID-GAN&Identity-Diversity Generative Adversarial Network &SSRC&Structured Sparse Representation Classification\\
InfoGAN&Information maximizing Generative Adversarial Network&SVM & Support Vector Machine \\
KCFA & Kernel Correlation Feature Analysis &VAE& Variational AutoEncoder\\ 
KLD-LGBP &Kullback-Leibler Divergence-Local Gabor Binary Patterns&YOLO & You Only Look Once\\
\bottomrule
\end{tabularx}
\end{table*}

\clearpage
\clearpage
\bibliographystyle{plain}
\newpage
\bibliography{references}

\end{document}